\documentclass[twoside,11pt]{article}

\usepackage{blindtext}

\usepackage[nohyperref]{jmlr2e}
\usepackage{amsmath}
\usepackage{amssymb}
\usepackage{mathtools}
\usepackage{microtype}
\usepackage{graphicx}
\usepackage{subfigure}
\usepackage{booktabs} 
\usepackage{bbm}
\usepackage{custom_commands}
\usepackage{bm}
\usepackage{enumitem}
\usepackage[noorphans,vskip=0ex]{quoting}
\usepackage{xcolor}
\usepackage{xspace}
\usepackage{etoolbox}
\usepackage{algorithm}
\usepackage{algorithmic}
\usepackage{hyperref}
\usepackage{multirow}
\definecolor{customcolor}{rgb}{	0.133, 0.545, 0.133}
\hypersetup{
    colorlinks=true,
    linkcolor=red,
    filecolor=magenta,      
    urlcolor=magenta,
    citecolor=blue,
    pdfpagemode=FullScreen,
    }
\usepackage[capitalize,noabbrev]{cleveref}

\usepackage{lastpage}
\jmlrheading{23}{2022}{1-\pageref{LastPage}}{10/23; Revised xx/xx}{xx/xx}{21-0000}{Haoxiang Wang, Gargi Balasubramaniam, Haozhe Si, Bo Li and Han Zhao}

\ShortHeadings{Invariant-Feature Subspace Recovery}{Wang, Balasubramaniam, Si, Li and Zhao}
\firstpageno{1}

\begin{document}

\title{Invariant-Feature Subspace Recovery: \\
A New Class of Provable Domain Generalization Algorithms}

\author{\name Haoxiang Wang$^\ast$ \email hwang264@illinois.edu \\
      \addr University of Illinois Urbana-Champaign
      \AND
      \name Gargi Balasubramaniam$^\ast$ \email gargib2@illinois.edu \\
      \addr University of Illinois Urbana-Champaign
      \AND
      \name Haozhe Si \email haozhes3@illinois.edu \\
      \addr University of Illinois Urbana-Champaign
      \AND
      \name Bo Li \email bol@uchicago.edu \\
      \addr University of Chicago
      \AND
      \name Han Zhao \email hanzhao@illinois.edu \\
      \addr University of Illinois Urbana-Champaign
      }

\newcommand{\customfootnotetext}[2]{{%
		\renewcommand{\thefootnote}{#1}%
		\footnotetext[0]{#2}}}%
\customfootnotetext{$\ast$}{Equal contribution.}

\editor{}

\maketitle

\begin{abstract}
Domain generalization asks for models trained over a set of training environments to generalize well in unseen test environments. Recently, a series of algorithms such as Invariant Risk Minimization (IRM) have been proposed for domain generalization. However, \citet{risks-of-IRM} shows that in a simple linear data model, even if non-convexity issues are ignored, IRM and its extensions cannot generalize to unseen environments with less than $d_s\mathrm{+}1$ training environments, where $d_s$ is the dimension of the spurious-feature subspace. In this work, we propose \textbf{I}nvariant-feature \textbf{S}ubspace \textbf{R}ecovery (ISR): a new class of algorithms to achieve provable domain generalization across the settings of classification and regression problems. First, in the binary classification setup of \citet{risks-of-IRM}, we show that our first algorithm, \textbf{ISR-Mean}, can identify the subspace spanned by invariant features from the first-order moments of the class-conditional distributions, and achieve provable domain generalization with $d_s\mathrm{+}1$ training environments. Our second algorithm, \textbf{ISR-Cov}, further reduces the required number of training environments to $\cO(1)$ using the information of second-order moments. Notably, unlike IRM, our algorithms bypass non-convexity issues and enjoy global convergence guarantees. Next, we extend ISR-Mean to the more general setting of multi-class classification and propose \textbf{ISR-Multiclass}, which leverages class information and provably recovers the invariant-feature subspace with $\lceil d_s/k \rceil + 1$ training environments for $k$-class classification. Finally, for regression problems, we propose \textbf{ISR-Regression} that can identify the invariant-feature subspace with $d_s + 1$ training environments. Empirically, we demonstrate the superior performance of our ISRs compared with IRM on synthetic benchmarks. Furthermore, ISRs can be used as simple yet effective post-processing methods for any given black-box feature extractors such as neural nets, and we show they can improve the worst-case accuracy of (pre-)trained models against spurious correlations and group shifts over multiple real-world datasets.

\end{abstract}


\section{Introduction}\label{sec:intro}
Domain generalization, i.e., out-of-distribution (OOD) generalization, aims to obtain models that can generalize to unseen (OOD) test domains after being trained on a limited number of training domains \citep{blanchard2011generalizing,wang2021generalizing,zhou2021domain,shen2021towards}. A series of works try to tackle this challenge by learning the so-called domain-invariant features (i.e., features whose distributions do not change across domains)~\citep{long2015learning,ganin2016domain,hoffman2018cycada,zhao2018adversarial,zhao2019learning,tachet2020domain}. On the other hand, Invariant Risk Minimization (IRM) \citep{IRM}, represents another approach that aims to learn features that induce invariant optimal predictors over training environments. Throughout this work, we shall use the term \emph{invariant features} to denote such features. There is a stream of follow-up works of IRM \citep{javed2020learning,REx,shi2020invariant,ahuja2020invariant,khezeli2021invariance,li2021learning,li2022invariant}, which propose alternative objectives or extends IRM to different settings as well.\newline
Recently, some theoretical works demonstrate that IRM and its variants fail to generalize to unseen environments, or cannot outperform empirical risk minimization (ERM), in various simple data models \citep{risks-of-IRM,kamath2021does,ahuja2021empirical}. For instance, \citet{risks-of-IRM} considers a simple Gaussian linear data model such that the class-conditional distribution of \textit{invariant features} remains the same across domains, while that of \textit{spurious features} changes across domains. Intuitively, a successful domain generalization algorithm is expected to learn an \textit{optimal invariant predictor}, which relies on only the invariant features and is optimal over the invariant features. To remove the noise introduced by finite samples, these theoretical works generally assume that infinite samples are available per training environment to disregard finite-sample effects, and the main evaluation metric for domain generalization algorithms is the \textit{number of training environments} needed to learn an optimal invariant predictor -- this metric is also referred to as \textit{environment complexity} in the literature~\citep{chen2021iterative}. In the case of linear predictors, \citet{risks-of-IRM} shows that IRM and REx (an alternative objective of IRM proposed in \citet{REx}) need $E>d_s$ to learn optimal invariant predictors, where $E$ is the number of training environments, and $d_s$ is the dimension of spurious features. In the case of non-linear predictors, they both fail to learn invariant predictors. Notice that the $E > d_s$ condition of IRM can be interpreted as a \textit{linear environment complexity} (i.e., $O(d_s)$ complexity), which is also observed in other recent works \citep{kamath2021does,ahuja2021empirical,chen2021iterative}. 
\begin{table*}[ht!]
\centering
    \resizebox{\textwidth}{!}{%
    \centering  
    \setlength{\tabcolsep}{4.5pt}
\begin{tabular}{cc|c|cc}
\toprule
\multirow{2}{*}{\textbf{Learning Paradigm} } &\multirow{2}{*}{ \textbf{Graphical Model}} & \multirow{2}{*}{\textbf{Algorithm}} &
\textbf{Provable Guarantee} & \multirow{2}{*}{\textbf{Theorem}} \\
& & & (\# Environments) & \\
\midrule
\multirow{2}{*}{Binary Classification} & \multirow{3}{*}{Anti-Causal ($y \rightarrow z_c$)} & \textsc{ISR-Mean} & $\cO(d_s)$ & Thm.\ \ref{thm:isr-mean}  \\
& & \textsc{ISR-Cov} & $\cO(1)$ & Thm.\ \ref{thm:isr-cov} \\
\cmidrule{3-5}
$k$-class Classification  & & \textsc{ISR-Multiclass} & $\cO(d_s / k)$ & Thm.\ \ref{thm:isr-multiclass} \\
\midrule
Regression  & Causal ($z_c \rightarrow y)$ & \textsc{ISR-Regression}& $\cO(d_s)$ & Thm.\ \ref{thm:isr-regression} \\
\bottomrule
\end{tabular}
}
\caption{Summary of our contributions.} 
\label{tab:contribution-summary}
\end{table*}
In this work, we propose a novel approach for domain generalization, Invariant-feature Subspace Recovery (ISR): which recovers the subspace spanned by invariant features, and then fits predictors in this subspace. More concretely, we present algorithms for provable recovery in the settings of both classification and regression problems with their corresponding environment complexities, as summarized in~\Cref{tab:contribution-summary}. We start with the simplest case of binary classification and present two algorithms to realize this approach, ISR-Mean and ISR-Cov, which utilize the first-order and second-order moments (i.e., mean and covariance) of class-conditional distributions, respectively. Under the linear data model of \citet{risks-of-IRM}, we prove that a) ISR-Mean is guaranteed to learn the optimal invariant predictor with $E\geq d_s+1$ environment, matching the environment complexity of IRM that is proved in~\citet{risks-of-IRM}, and b) ISR-Cov reduces the requirement to $E\geq 2$, achieving a constant $O(1)$ environment complexity. Notably, both ISR-Mean and ISR-Cov require fewer assumptions on the data model than IRM, and they both enjoy global convergence guarantees, while IRM does not because of the non-convex formulation of its objective. Furthermore, the ISRs are also more computationally efficient than IRM, since the computation of ISRs involves basically only empirical risk minimization (ERM) with one additional call of an eigen-decomposition solver. Next, we extend ISR-Mean to the setting of multi-class classification and propose ISR-Multiclass, which leverages information from multiple classes to provably recover the invariant-feature subspace from $E\geq \lceil d_s/k \rceil + 1$ environments for $k$-class classification. We then consider the setting of regression and present ISR-Regression, which provably recovers the invariant-feature subspace in $E\geq d_s+1$ environments. 

Empirically, we conduct studies on a set of challenging synthetic linear benchmarks adapted from \cite{aubin2021linear}, semi-synthetic image benchmarks with strong spurious correlations (variants of Colored MNIST), and a suite of real-world datasets (two image datasets and one text dataset used in \citet{sagawa2019distributionally}, with one additional tabular dataset). Our results on the synthetic benchmarks empirically validate our proved environment complexities, and also demonstrate its superior performance when compared with IRM and its variants. Since the real-world data are highly complex and non-linear, over which the ISR approach cannot be directly applied, we apply ISR on top of the features extracted by the hidden layers of trained neural nets as a post-processing procedure. Experiments show that ISR can consistently increase the worse-case accuracy of the trained models against spurious correlations and group shifts, and this includes models trained by ERM, IRM, Information Bottleneck, reweighting, MixUp, and GroupDRO~\citep{sagawa2019distributionally}.

\section{Related Work}\label{sec:related-works}

\textbf{Domain Generalization.}~Domain generalization (DG), also known as OOD generalization, aims at leveraging the labeled data from a limited number of training environments to improve the performance of learning models in unseen test environments~\citep{blanchard2011generalizing,muandet2013domain}. The simplest approach for DG is empirical risk minimization i.e.\ ERM~\citep{vapnik1992principles}, which minimizes the sum of empirical risks over all training environments. Distributionally robust optimization is another approach \citep{sagawa2019distributionally,volpi2018generalizing}, which optimizes models over a worst-case distribution that is perturbed around the original distribution. Besides, there are two popular approaches, domain-invariant representation learning and invariant risk minimization, which we will discuss in detail below. In addition to algorithms, there are works that propose theoretical frameworks for DG \citep{zhang2021quantifying,ye2021towards}, or empirically examine DG algorithms over various benchmarks \citep{gulrajani2021in,koh2021wilds,wiles2021fine}. 
Notably, some recent works consider DG with temporarily shifted environments \citep{koh2021wilds,ye2022future,wang2022understanding}, which is a novel and challenging setting. Besides DG, there are other learning paradigms that involve multiple environments, such as multi-task learning \citep{caruana1997multitask,wang2021bridging} and meta-learning \citep{finn2017model,wang2022global}, which do not aim at generalization to OOD environments. 

\paragraph{Domain-Invariant Representation Learning.}~Domain-Invariant representation learning is a learning paradigm widely applied in various tasks. In particular, in domain adaptation (DA), many works aim to learn a representation of data that has an invariant distribution over the source and target domains, adopting methods including adversarial training \citep{ganin2016domain,tzeng2017adversarial,zhao2018adversarial} and distribution matching \citep{ben2007analysis,long2015learning,sun2016deep}. The domain-invariant representation approach for DA enjoys theoretical guarantees~\citep{ben2010theory}, but it is also pointed out that issues such as conditional shift should be carefully addressed~\citep{zhao2019learning}. In domain generalization~\citep{blanchard2011generalizing}, since there is no test data (even unlabelled ones) available, models are optimized to learn representations invariant over training environments~\citep{albuquerque2020generalizing,chen2021iterative}. Notice that many domain-invariant representation learning methods for DA can be easily applied to DG as well~\citep{gulrajani2021in}.
\paragraph{Invariant Risk Minimization.}~\citet{IRM} proposes invariant risk minimization (IRM) that aims to learn invariant predictors over training environments by optimizing a highly non-convex bi-level objective. The authors also reduce the optimization difficulty of IRM by proposing a practical version, IRMv1, with a penalty regularized objective instead of a bi-level one. Alternatives of IRM have also been studied \citep{ahuja2020invariant,li2022invariant}. However, \citet{risks-of-IRM,kamath2021does,ahuja2021empirical} theoretically show that these algorithms fail even in simple data models.

\paragraph{Spurious Correlation Mitigation}
In this work, the focus is on a specific kind of distribution shifts involving robustness to \textit{spurious correlations}: a model should not rely on features that might appear to be \textit{spuriously} correlated with the target variable in certain domains. One would like to learn a model which is robust to these spurious correlations and performs well on \textit{all} subpopulations of data during test time~\citep{sagawa2019distributionally}, even where the spurious correlations break. Recent work throws light on the finding that ERM is able to learn both ``core'' and ``spurious'' features where appropriate linear probing~\citep{kumar2022fine} may be sufficient to find a good predictor. For example,~\citet{kirichenko2022last} demonstrate that re-training the last layer with access to a balanced validation dataset or ``deep feature weighting'' matches or outperforms state-of-the-art methods like GroupDRO \citep{sagawa2019distributionally}, which explicitly minimizes the worst group loss by leveraging domain membership of every sample. Subsequently,~\citet{lee2022surgical} propose ``surgical fine-tuning'', claiming that only fine-tuning the last layer for spurious correlation problems performs better than fine-tuning the entire network. This is part of a broader claim that one should fine-tune only those network parameters which are responsible for the observed distribution shift. Another approach involves contrastive representation learning~\citep{zhang2022correct} for improved robustness to spurious correlations in the absence of domain-specific labels. In the setting of regression,~\citet{rosenfeld2022domain} make similar claims on ERM learning and perform a domain-specific transformation such that different domains share similar optimal predictors. Other perspectives include~\citet{yao2022c} which introduces C-MixUp for regression, a technique that employs sampling of input points to be linearly interpolated as a data augmentation strategy for better in-domain and out-of-domain generalization. In a parallel work,~\citet{ahuja2021invariance} and~\citet{li2022invariant} propose an information bottleneck regularizer in the form of a variance penalty on the feature representations to the original ERM and IRM objectives in the learning process, demonstrating improved OOD performance. Note that both C-MixUp and information bottleneck require end-to-end training of the model, as opposed to the computationally efficient post-processing approach introduced in this work.

\begin{figure}[htb]
\begin{center}
\centerline{\includegraphics[width=.5\columnwidth]{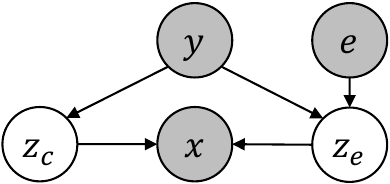}}
\caption{The causal graph of the data model in \citet{risks-of-IRM}. Shading represents that the variable is observed.
}
\label{fig:cgraph}
\end{center}
\end{figure}

\section{Problem Setup}\label{sec:setup}
\paragraph{Notation}
Each labeled example can be represented as a $(x,y,e)$ tuple, where $x\in \bR^d$ is the input, $y$ is the label which can be discrete or continuous, and $e\in \mathbb{Z}_+$ is the index of the environment that provides $(x,y)$. In addition, we assume $x$ is generated by a latent feature $z\in \bR^d$, which generates $x$ and is correlated with $y$ and $e$  (e.g., see the example in Fig.~\ref{fig:cgraph}). Besides, we use $X,Y,\mathscr E, Z$ to refer to random variables w.r.t. $x,y,e,z$. 

In this paper, we first adopt the linear Gaussian data model of~\citet{risks-of-IRM} for the case of binary classification in~\Cref{sec:isr-mean} and then extend it to the setting of multi-class classification in~\Cref{sec:isr-multiclass}. We also discuss a new causal model for regression in~\Cref{sec:isr-regression}.

\subsection{Binary Classification} \label{sec:setup-binary}
As per the linear Gaussian data model of~\citet{risks-of-IRM}, it is assumed that the training data are drawn from $E$ training environments, $\mathcal E = \{1,..., E\}$. For arbitrary training environment $e\in \mathcal E$, each sample in this environment is generated by the following mechanism (see Fig. \ref{fig:cgraph} for an illustration): first, a label $y\in \{\pm 1\}$ is sampled,
\begin{align} \label{eq:def-y}
    y & = \begin{cases}
    1, & \text{with probability } \eta\\
    -1, & \text{otherwise}
    \end{cases}
\end{align}
Then, both invariant latent features $z_c$ and spurious latent features $z_e$ of this sample are drawn from the following Gaussian distributions:
\begin{align}\label{eq:def-latent-features}
    z_c \sim \mathcal{N}(y\mu_c, \sigma_c^2 I) \in \mathbb{R}^{d_c}, 
    z_e \sim \mathcal{N}(y  \mu_e, \sigma_e^2 I) \in \mathbb{R}^{d_s}  
\end{align}
where $\mu_c \in \mathbb{R}^{d_c}, \mu_e \in \mathbb{R}^{d_s}$ and $\sigma_c,\sigma_e \in \mathbb{R}_+$. The constants $d_c$ and $d_s$ refer to the dimension of invariant features and spurious features, respectively. The total number of feature attributes is then $d=d_c+d_s$. Notice that $\mu_c,\sigma_c$ are invariant across environments, while $\mu_e,\sigma_e$ are dependent on the environment index $e$. Following \citet{risks-of-IRM}, we name $\{\mu_e\}$ and $\{\sigma_e\}$ as \textit{environmental} means and variances.

\citet{risks-of-IRM} adopts a mild non-degeneracy assumption\footnote{It was stated as (9) in \citet{risks-of-IRM}.} on the environmental mean from the IRM paper \citep{IRM}, stated as Assumption \ref{assum:non-degenerate-mean} below. In addition, the authors also make another non-degeneracy assumption\footnote{It is stated as Eq. (10) in \citet{risks-of-IRM}, which is a sufficient (not necessary) condition for our Assumption \ref{assum:non-degenerate-cov}.} on the environmental variances, which we relax to the following Assumption \ref{assum:non-degenerate-cov}.
\begin{assumption}\label{assum:non-degenerate-mean} The set of environmental means $\{\mu_e\}_{e=1}^E$ is affinely independent. 
\end{assumption}
\begin{assumption}\label{assum:non-degenerate-cov}
Assume there exists a pair of distinct training environments $e,e'\in[E]$ such that $\sigma_e\neq \sigma_{e'}$.
\end{assumption}
With the latent feature $z$ as a concatenation of $z_c$ and $z_e$, the observed sample $x$ is generated by a linear transformation on this latent feature. For simplicity, we consider that $x$ has the same dimension as $z$.
\begin{align}\label{eq:def-linear-transform}
    z = \begin{bmatrix}
    z_c\\
    z_e
    \end{bmatrix} \in \bR^{d}, \quad 
    x = R z = A z_c + B z_e \in \bR^d
\end{align}
where $d= d_c + d_s$, and $A = \bR^{d \times d_c}, B = \bR^{d \times d_s}$ are fixed transformation matrices with concatenation as $R = [A,B]\in \bR^{d\times d}$. Then, each observed sample $x$ is effectively a sample drawn from 
\begin{align}\label{eq:transformed-gaussian}
    \mathcal N(y(A \mu_c + B \mu_e), \sigma_c^2 AA^\T + \sigma_e^2 BB^\T)
\end{align}
The following assumption is also imposed on the transformation matrix in~\citet{risks-of-IRM}:
\begin{assumption}\label{assum:full-rank-transform} $R$ is injective.
\end{assumption}
Since $R\in \bR^{d \times d}$, Assumption \ref{assum:full-rank-transform} leads to the fact $\mathrm{rank}(R) = d$, indicating that $R$ is full-rank.

\subsection{Multi-Class Classification}\label{sec:setup-multiclass}
Consider a $k$ class classification problem under the causal graph in Fig.~\ref{fig:cgraph}. Let $y$ be sampled from a prior distribution of labels $\{y_1, y_2, \dots, y_k\}$. Now, the invariant features $z_c \in \bR^{d_c}$ and spurious features $z_e \in \bR^{d_s}$ are sampled as follows:
\begin{align}
    z_c \sim \mathcal{N}(\mu_{y}, \sigma_c^2 I_{d_c}) 
    , 
    z_e \sim \mathcal{N}(\mu_{ye}, \sigma_e^2 I_{d_s}) 
\end{align} where $\mu_y \in \mathbb{R}^{d_c}, \mu_{ye} \in \mathbb{R}^{d_s}$ and $\sigma_c,\sigma_e \in \mathbb{R}_+$. 

Note that in the binary classification data model discussed in section \ref{sec:setup-binary}, the invariant feature distribution is given by $z_c\sim \mathcal{N}(y\mu_{c}, \sigma_c^2 I_{d_c})$, where $y\in\{+1, -1\}$. Here, $y$ induces \textit{symmetric means} for the invariant (and similarly spurious features) i.e.\ $\pm \mu_c$ ($\pm \mu_e$). We will see in section \ref{sec:isr-multiclass} that this symmetry assumption is \textit{unnecessary} and artificially increases the environment complexity from $\lceil d_s / 2 \rceil + 1$ to $d_s + 1$. Thus, we consider a more general and flexible causal model where the means of invariant (and spurious features) depend on the specific value of $y$.

Finally, the input $x$ is generated from a linear transformation of the concatenated invariant and spurious features as per steps~\eqref{eq:def-linear-transform} and~\eqref{eq:transformed-gaussian}. We also adopt the full-rank assumption on the means of the spurious features from different environments and different classes.
\begin{assumption}
\label{assum:non-degenerate-mean-multiclass} 
For the set of environmental means, $\{\mu_{ye}\}_{e=1}^E,_{y=1}^k$, we assume that 
\begin{align}
    \operatorname{dim(span}(\{\mu_{ye}: y\in[k],e\in[E]\})) = \operatorname{min}(E \times k, d_s)
\end{align}
\end{assumption}

Intuitively, this assumption ensures that multiple classes / multiple environments do not \textit{trivially replicate} information, and thus the spurious (and thus invariant) subspace can be recovered. When $k = 1$, this assumption is equivalent to Assumption \ref{assum:non-degenerate-mean}.

Finally, we also assume that~\eqref{assum:full-rank-transform} holds, i.e., $R$ is injective. 

\subsection{Regression}\label{sec:setup-regression}

\begin{figure}[h]
    \centering
    \includegraphics[scale=0.7]{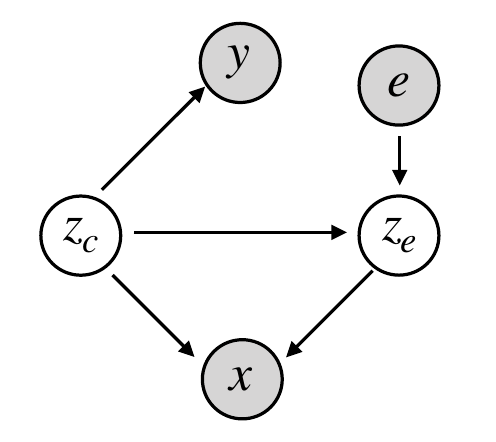}
    \caption{Causal Graph for Regression. Shaded variables are observed.}
    \label{fig:cgraph-regression}
\end{figure}

In the setting of regression, $y$ is a continuous-valued attribute i.e.\ $y \in \bR$. We propose the causal graph\footnote{Note that~\citet{ahuja2021invariance} assume a similar causal model with the difference that all nodes in the graph depend on the environment index $e$.} as shown in Figure \ref{fig:cgraph-regression}. 

First, the invariant feature $z_c$ is sampled as follows:
\begin{align}
    z_c &\sim \mathcal{N}(\mu_{c}, \sigma_c^2 I_{d_c}) 
\end{align}
where $\mu_c \in \mathbb{R}^{d_c}$,  $\sigma_c \in \mathbb{R}_+$. The target $y$ is then determined by a function of $z_c$:
\begin{align}
    y = f(z_c) = w_c^\top z_c + b_c \in \bR
\end{align}
Further, the spurious features $z_e$ depend on both the invariant features and the environment index $e$, which induces a spurious correlation between $y$ and $z_c$. Specifically,
\begin{gather}\label{eq:zszc-regression}
    z_e = W_{cs}^ez_c + b_e \in \bR^{d_s},~\quad W_{cs}^e \in \mathbb{R}^{d_s \times d_c}
\end{gather}
Note that $W_{cs}^e$ is an \textit{environment specific} transformation for obtaining the spurious features. Finally, the input $x$ is generated from a linear transformation of the concatenated invariant and spurious features as follows:

\begin{align}\label{eq:def-linear-transform-regression}
    z = \begin{bmatrix}
    z_c\\
    z_e
    \end{bmatrix} \in \bR^{d}, \quad 
    x = R z = A z_c + B z_e \in \bR^d
\end{align}

The following assumptions is made in this setting:
\begin{assumption}\label{assum:non-degenerate-wcs-regression} 
For every environment, $e$, $W_{cs}^e$ is full rank, and $d_s \leq d_c$.
\end{assumption}

This non-degeneracy assumption ensures that the spurious features $z_s$ have full dimension $d_s$ - since we obtain $z_s$ from $z_c$ via the transformation described in \eqref{eq:zszc-regression}, the effective dimension of $z_s$ i.e.\ $d_s$ cannot be greater than that of the input space i.e., $d_c$.\footnote{Note that our proposed method still holds without the assumption that $d_s\leq d_c$. If $W_{cs}^e$ is not full rank, then the effective dimension of $z_s$ will be smaller than $d_s$. However, our theorem requires $E > d_s \geq \rank(W_{cs}^e)$, which still holds.}.

Similar to the classification setting, we also make the following assumption on the set of environmental means and the linear map $R$:
\begin{assumption}
    The set of environmental means $\{\mu_e\}_{e=1}^E$ is affinely independent and $R$ is injective.
\end{assumption}

\subsection{Optimal Invariant Predictors and IRM}

Denote the data of any training domain $e$ as $\mathcal D_e$. During training, learners have access to the environment index $e$ for each training sample, i.e., learners observe samples in the form of $(x,y,e)$.

\paragraph{Optimal Invariant Predictors}
The quest of IRM is to find the optimal invariant predictors, i.e., classifiers/regressors that use only invariant features and are optimal w.r.t. invariant features over the training data. In the data model of \citet{risks-of-IRM}, because of the linear nature of the data generation process, the optimal invariant predictors are contained in the linear function class. If the task of consideration is binary classification, \citet{risks-of-IRM} chooses the logistic loss as the loss function for optimization\footnote{\citet{risks-of-IRM} proves that logistic loss over linear models can attain Bayes optimal classifiers in this data model.}, which we also adopt in this work.
Then, the goal of domain generalization is to learn a linear featurizer (feature extractor) $\Phi$ and a linear classifier $ \beta$ that minimizes the risk (population loss) on any unseen environment $e$ with data distribution $p_e$ satisfying Assumptions~\eqref{eq:def-y}-\eqref{eq:def-linear-transform}:
\begin{align}
    \mathcal{R}^{e}(\Phi, \beta):=\mathbb{E}_{(x, y) \sim p^{e}}\left[\ell\left(w^{\T} \Phi(x) + b ,~ y\right)\right]
\end{align}
where $\ell$ is logistic loss (for binary classification), cross-entropy loss (for multi-class classification) or squared loss (for regression), and $\beta=(w,b)$ with weight $w$ and bias $b$.  

To be complete, we present the optimal invariant predictor derived by \citet{risks-of-IRM} in the setting of binary classification as follows:
\begin{proposition}[Optimal Invariant Predictor]\label{prop:optimal-inv-pred}
Under the data model considered in Eq.~\eqref{eq:def-y}-\eqref{eq:def-linear-transform}, the optimal invariant predictor $h^*$ w.r.t.\ logistic loss is unique, which can be expressed as a composition of i) a featurizer $\Phi^*$ that recovers the invariant features and ii) the classifier $\beta^*=(w^*,b^*)$ that is optimal w.r.t. the extracted features:
\begin{align}
    h^*(x) &= {w^*}^\T \Phi^*(x) + b^*\label{eq:optimal-inv-pred}\\
    \Phi^*(x)&\coloneqq
    \begin{bmatrix}
    I_{d_c} & 0\\
    0 & 0
    \end{bmatrix} R^{-1} x= 
    \begin{bmatrix}
    z_c\\
    0
    \end{bmatrix}\in \bR^{d \times d}\\
    w^*&\coloneqq 
    \begin{bmatrix}
    2 \mu_c / \sigma_c^2\\
    0
    \end{bmatrix}\in \bR^{d}, \quad 
    b^* \coloneqq \log \frac{\eta}{1-\eta} \in \bR
\end{align}
\end{proposition}
Notice that even though the optimal invariant predictor $h^*$ is unique, its components (the featurizer and classifier) are only unique up to invertible transformations. For instance, $( {w^*}^\T U^{-1})(U \Phi) = {w^*}^\T \Phi $ for any invertible $U\in \bR^{d\times d}$.

\paragraph{Invariant Risk Minimization}
IRM optimizes a bi-level objective over a featurizer $\Phi$ and a classifier $\beta$,
\begin{align}\label{eq:IRM}
    \mathrm{IRM:}~~&\min_{\Phi,\beta} \sum_{e\in [E]} \mathcal{R}^{e}(\Phi, \beta) \\
    &\mathrm{s.t.}~ \beta \in \argmin_{\beta}\mathcal{R}^e(\Phi, \beta) ~~\forall e\in[E]\nonumber
\end{align}
This objective is non-convex and difficult to optimize. Thus, \citet{IRM} proposes a Langrangian form to find an approximate solution, 
\begin{align}\label{eq:IRMv1}
    \mathrm{IRMv1:}~ \min _{\Phi, \hat{\beta}} \sum_{e \in [E]}\mathcal{R}^{e}(\Phi, \hat{\beta})+\lambda\left\|\nabla_{\hat{\beta}} \mathcal{R}^{e}(\Phi, \hat{\beta})\right\|_{2}^{2}
\end{align}
where $\lambda > 0$ controls the regularization strength. Notice that the IRMv1$\eqref{eq:IRMv1}$ is still non-convex, and it becomes equivalent to the original IRM \eqref{eq:IRM} as $\lambda \rightarrow \infty$.

\paragraph{Environment Complexity}
To study the dependency of domain generalization algorithms on environments, recent theoretical works \citep{risks-of-IRM,kamath2021does,ahuja2021invariance,chen2021iterative} consider the ideal setting of infinite data per training environment to remove the finite-sample effects. In this infinite-sample setting, a core measure of domain generalization algorithms is \textit{environment complexity}: the number of training environments needed to learn an invariant optimal predictor. For this data model, \citet{risks-of-IRM} proves that with the linear $\Phi$ and $\beta$, the environment complexity of IRM is $d_s+1$, assuming the highly non-convex objective \eqref{eq:IRM} is optimized to reach the global optimum. This linear environment complexity (i.e., $\cO(d_s)$) of IRM is also proved in \citep{kamath2021does,ahuja2021invariance} under different simple data models.

\setlength{\textfloatsep}{10pt}
\begin{algorithm}[tb]
\caption{ISR-Mean}\label{algo:isr-mean}
\begin{algorithmic}
\STATE {\bfseries Input:} Data of all training environments, $\{\mathcal D_e\}_{e\in [E]}$.
\FOR{$e = 1,2,\dots,E$}     
    \STATE Estimate the sample mean of $\{x|(x,y)\in \mathcal D_e, y=1\}$ as $\bar x_e\in \bR^{d}$
\ENDFOR
\STATE \textbf{I.} Construct a matrix $\mathcal M \in \bR^{E \times d}$ with the $e$-th row as $\bar{x}_e^\T$ for $e\in [E]$
\STATE \textbf{II.} Apply PCA to $\mathcal M$ to obtain eigenvectors $\{P_1,...,P_d\}$ with eigenvalues $\{\lambda_1,...,\lambda_d\}$
\STATE \textbf{III.} Choose $d_s$ eigenvectors corresponding to the highest eigenvalues and stack them to obtain a transformation matrix $P''\in \bR^{d_s\times d}$. Take the null space of $P''$ to obtain transformation matrix $P'\in \bR^{d_c\times d}$
\STATE \textbf{IV.} Fit a linear classifier (with $w\in \bR^{d_c}$, $b\in \bR$) by ERM over all the training data with transformation $x\mapsto P'x$
\STATE Obtain a predictor $f(x) =  w^\T P' x + b$
\end{algorithmic}
\end{algorithm}

\section{Invariant-Feature Subspace Recovery}\label{sec:algo}

In this section, we introduce four algorithms for invariant-feature subspace recovery in the setting of binary classification (\ref{sec:isr-binary}), multi-class classification (\ref{sec:isr-multiclass})and regression(\ref{sec:isr-regression}). We start with binary classification, where we present ISR-Mean(\ref{sec:isr-mean}) and ISR-Cov(\ref{sec:isr-cov}), which recover the invariant-feature subspace with the first-order and second-order moments of class-conditional data distributions, respectively. We then present extensions of ISR-Mean to the setting of multi-class classification (ISR-Multiclass, \ref{sec:isr-multiclass}) and regression (ISR-Regression, \ref{sec:isr-regression}).

\subsection{Binary Classification}\label{sec:isr-binary}

\subsubsection{ISR-Mean} \label{sec:isr-mean}
Algorithm \ref{algo:isr-mean} shows the pseudo-code of ISR-Mean, and we explain its four main steps in detail below. In the setup of \cref{sec:setup}, ISR-Mean enjoys a linear environment complexity that matches that of IRM, while requiring fewer assumptions (no need for Assumption \ref{assum:non-degenerate-cov}).

\paragraph{I. Estimate Sample Means across Environments}
In any training environment $e$, each observed sample $x\in \bR^d$ is effectively drawn i.i.d. from $\mathcal N (y (A \mu_c + B \mu_e), A A^\T \sigma_c^2 + B B^\T \sigma_e^2)$, as stated in \eqref{eq:transformed-gaussian}. In the infinite-sample setting considered in \cref{sec:setup}, the mean of the positive-class data in environment $e$ can be expressed as $\E[X|Y=1, \mathscr E =e]$, which is exactly the value of $\bar{x}_e$ in \cref{algo:isr-mean}. Thus, we know $\bar x_e$ satisfies $\bar{x}_e = A \mu_c + B \mu_e$, and the matrix $\mathcal M$ can be expressed as
\begin{align}\label{eq:def-M}
    \mathcal M \mathrm{\coloneqq} \begin{bmatrix}
    \bar{x}_1^\T\\
    \vdots\\
    \bar{x}_E^\T
    \end{bmatrix} \mathrm{=} \begin{bmatrix}
    \mu_c^\T A^\T \mathrm{+} \mu_1^\T B^\T\\
    \vdots\\
    \mu_c^\T A^\T \mathrm{+} \mu_E^\T B^\T\end{bmatrix}
    \mathrm{=}{\overbrace{\begin{bmatrix}
    \mu_c^\T ~~\mu_1^\T\\
    \vdots~~~~~\vdots\\
    \mu_c^\T ~~ \mu_E^\T
    \end{bmatrix}}^{\mathcal U^\T \coloneqq}} R^\T
\end{align}

\begin{figure}[tb]
\begin{center}
\centerline{\includegraphics[width=.5\columnwidth]{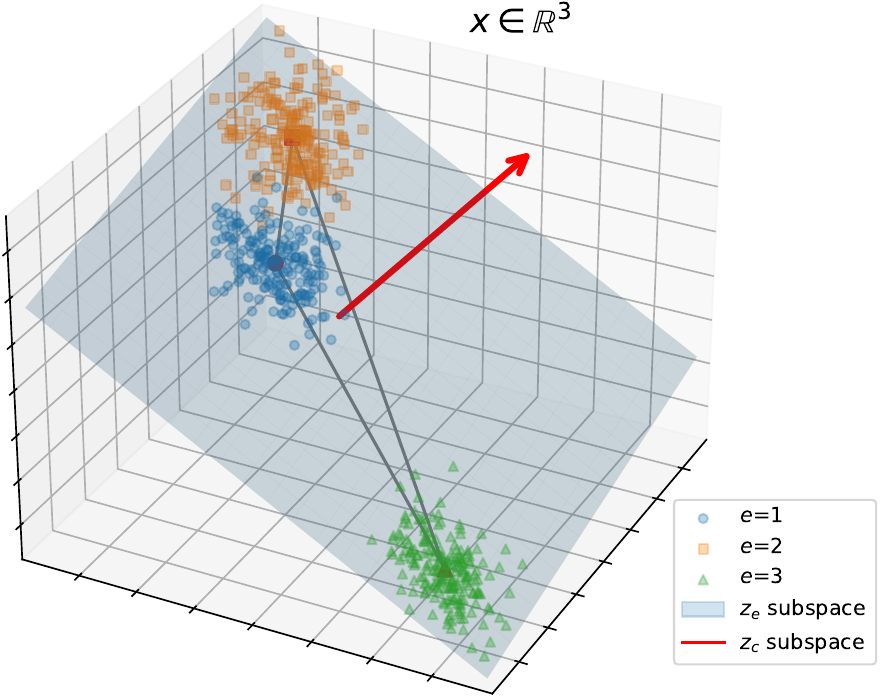}}
\caption{An example for ISR-Mean with $d_c\mathrm{=}1$, $d_s\mathrm{=}2$, $E\mathrm{=}3$. In this $\bR^3$ input space, the blue 2D plane is determined by sample means of positive-class samples of the $3$ training environments. 
}
\label{fig:ISR-mean-demo}
\end{center}
\end{figure}

\paragraph{II. PCA on $\mathcal M$.}
In this step, we apply principal component analysis (PCA) \citep{pearson1901PCA} to the matrix $\mathcal M$. First, PCA performs mean-substraction on $\mathcal M$ to shift the sample mean of each column to zero, and we denote the shifted matrix as $\wt{\mathcal M}$. Then, PCA eigen-decompose $\wh \Sigma_\cM \coloneqq \frac{1}{E} \wt{\cM}^\T \wt{\cM}$, the sample covariance matrix of $\wt\cM$, such that $\wh \Sigma_\cM = P^\T S P$, where $P = [P_1,\dots,P_d]\in \bR^{d\times d}$ is a stack of eigenvectors $\{P_i\}_{i\in d}^d$, and $S\in \bR^{d \times d}$ is a diagonal square matrix with diagonal entries as eigenvalues $\{\lambda_i\}_{i=1}^d$ of $\wt \Sigma_{\cM}$. We consider the eigenvalues $\{\lambda_i\}_{i=1}^d$ are sorted in ascending order.

\paragraph{III: Recover the Invariant-Feature Subspace} 
Stack the eigenvectors corresponding to the highest $d_s$ eigenvalues to obtain a transformation matrix $P'' \in \bR^{d_s \times d}$. Then, take the null space of this matrix to obtain $P' \in \bR^{d_c \times d}$. 

Note that when $E \geq d_s + 1$, one can choose the \textit{lowest} $d_c$ eigenvalues and stack the corresponding eigenvectors to obtain a transformation directly into the invariant-feature subspace. However, when $E < d_s + 1$, this will not be equivalent to the presented algorithm, a discussion of which is deferred to section \ref{supp:proof:isr-mean}.

\paragraph{IV. Train a Classifier in the Invariant-Feature Subspace}
In this final step, we just transform all the training data by the transformation $x\mapsto P'x$, and fit a linear classifier with ERM to the transformed data to obtain an predictor,
\begin{align}\label{eq:ISR-Mean-predictor}
    f(x) = w^\T P' x + b
\end{align}
which is guaranteed to be the optimal invariant predictor $h^*$ defined in \cref{prop:optimal-inv-pred}, i.e., $f\equiv h^*$.

\paragraph{Global Convergence Guarantee}
ISR-Mean is guaranteed to converge to a global optimum since a) the step I and III are optimization-free, b) PCA can be efficiently solved to global the optimum by various methods \citep{arora2012stochastic,vu2013fantope,hauser2018pca,eftekhari2020principal}, c) the ERM objective of linear classifiers with logistic loss is convex, enjoying global convergence. 

\paragraph{Geometric Interpretation}
We provide a geometric interpretation of ISR-Mean with a 3D example in Fig.~\ref{fig:ISR-mean-demo}, where $d_c\mathrm{=}1$, $d_s\mathrm{=}2$, $E\mathrm{=}3$. For each environment $e$, the sample mean of its positive-class data, $\bar{x}_e$, must lie in a $d_s$-dimensional spurious-feature subspace in the infinite-sample setting, as proved by \cref{thm:isr-mean}. ISR-Mean aims to identify this spurious-feature subspace, and take its tangent subspace as the invariant-feature subspace.

\paragraph{Linear Environment Complexity}
In the infinite-sample setting, we prove below that with more than $d_s$ training environments, ISR-Mean is guaranteed to learn the invariant optimal predictor (\cref{thm:isr-mean}). Notice that even though this linear environment complexity is identical to that of IRM (proved in Theorem 5.1 of \citet{risks-of-IRM}), our ISR-Mean has two additional advantages: (a) Unlike IRM, ISR-Mean does not require any assumption on the covariance\footnote{IRM needs a covariance assumption stronger than our Assumption \ref{assum:non-degenerate-cov}, as pointed out in Sec. \ref{sec:setup}.} such as Assumption \ref{assum:non-degenerate-cov}; (b) ISR-Mean enjoys the global convergence guarantee, while IRM does not due to its non-convex formulation. The proof is in \cref{supp:proof:isr-mean}.

\begin{theorem}[ISR-Mean]\label{thm:isr-mean}
Suppose $E > d_s$ and the data size of each environment is infinite, i.e., $|\mathcal D_e| \mathrm{\rightarrow} \infty$ for $e\mathrm{=}1,\dots,E$. For PCA on the $\cM$ defined in \eqref{eq:def-M}, the obtained eigenvectors $\{P_1,\dots, P_d\}$ with corresponding ascendingly ordered eigenvalues $\{\lambda_1,\dots, \lambda_d\}$ satisfy
\begin{align*}
    \forall 1\leq i \leq d_c, ~\lambda_i = 0 \quad \text{and} \quad  \forall d_c < i \leq d, ~\lambda_i > 0
\end{align*}
The eigenvectors corresponding to these zero eigenvalues, i.e., $\{P_1,\dots, P_{d_c}\}$, can recover the subspace spanned by the invariant latent feature dimensions, i.e.,
\begin{align}\label{eq:thm:ISR-Mean:span-equality}
\mathrm{Span}(\{P_1^\T R,\dots, P_{d_c}^\T R\}) = \mathrm{Span}(\{\mathbf{\hat d_c^1},\dots, \mathbf{\hat d_c^{d_c}}\})
\end{align}
where $\mathbf{\hat d_c^i}$ is the unit-vector along the $i$-th coordinate in the latent feature space for $i=1,\dots, d_c$. Then, the classifier $f$ fitted with ERM to training data transformed by $x\mapsto [P_1,\dots,P_{d_c}]^\T x$ is guaranteed to be the invariant optimal predictor, i.e., $f = h^*$, where $h^*$ is defined in \eqref{eq:optimal-inv-pred}.
\end{theorem} 
\subsubsection{ISR-Cov} \label{sec:isr-cov}
The pseudo-code of ISR-Cov is presented in Algorithm \ref{algo:isr-cov}, with a detailed explanation below.
In the setup of \cref{sec:setup}, ISR-Cov attains an $\cO(1)$ environment complexity, the optimal complexity any algorithm can hope for, while requiring fewer assumptions than IRM (no need for Assumption \ref{assum:non-degenerate-mean}).

\paragraph{I. Estimate and Select Sample Covariances across Environments}
As~Eq.\eqref{eq:transformed-gaussian} indicates, in any environment $e$, each observed sample $x\in \bR^d$ with $y=1$ is effectively drawn i.i.d. from $\mathcal N (A \mu_c + B \mu_e, A A^\T \sigma_c^2 + B B^\T \sigma_e^2)$. Thus, the covariance of the positive-class data in environment $e$ can be expressed as $\mathrm{Cov}[X | Y=1, \mathscr{E}=e] = A A^\T \sigma_c^2 + B B^\T \sigma_e^2$, which is the value that $\Sigma_e$ in step I of \cref{algo:isr-cov} estimates. The estimation is exact in the infinite-sample setting of consideration, so we have $\Sigma_e = A A^\T \sigma_c^2 + B B^\T \sigma_e^2$. Assumption \ref{assum:non-degenerate-cov} guarantees that we can select a pair of environments $e_1,e_2$ with $\Sigma_1 \neq \Sigma_2$. Then, we have
\begin{align}\label{eq:delta-sigma-expression}
\Delta \Sigma \coloneqq \Sigma_{e_1} - \Sigma_{e_2} = (\sigma_{e_1}^2 - \sigma_{e_2}^2)BB^\T \in \bR^{d \times d}
\end{align}

\paragraph{II. Eigen-decompose $\Delta \Sigma$} 
Similar to the step II of \cref{algo:isr-mean} explained in Sec. \ref{sec:isr-mean}, we eigen-decompose $\Delta\Sigma$ to obtain eigenvectors $\{P_i\}_{i=1}^d$ 
corresponding to eigenvalues $\{\lambda_i\}_{i=1}^d$. We consider the eigenvalues are sorted in ascending order by their \textit{absolute values}.

\paragraph{III. Recover the Invariant-Feature Subspace}
As we shall formally prove in \cref{thm:isr-cov}, in the infinite-sample setting, a) the eigenvalues $\{\lambda_i\}_{i=1}^d$ 
should exhibit a ``phase transition'' phenomenon such that the first $d_c$ eigenvalues all are \textit{zeros} while the rest are all \textit{non-zero}, b) the $d_c$ eigenvectors corresponding to zero eigenvalues, $\{P_1,\dots,P_{d_c}\}$, are guaranteed to recover the $d_c$-dimensional invariant-feature subspace. We stack these eigenvectors as a matrix $P'$
\begin{align}\label{eq:trans-matrix-isr-cov}
    P' \coloneqq [P_1,\dots, P_{d_c}]^\T \in \bR^{d_c \times d}
\end{align}

\paragraph{IV. Train a Classifier in the Invariant-Feature Subspace}
This final step is the same as the step IV of Algorithm \ref{algo:isr-mean} described in Sec.~\ref{sec:isr-mean}.

\paragraph{Global Convergence}
Applying the same argument in Sec.~\ref{sec:isr-mean}, it is clear that ISR-Cov also enjoys the global convergence guarantee: the eigen-decomposition and ERM can both be globally optimized.

\paragraph{Improving the Robustness of ISR-Cov}
In practice with finite data, Algorithm~\ref{algo:isr-cov} may be non-robust as $\sigma_e$ and $\sigma_{e'}$ become close to each other. The noise due to finite samples could obfuscate the non-zero eigenvalues so that they are indistinguishable from the zero eigenvalues. To mitigate such issues, we propose a robust version of ISR-Cov that utilizes more pairs of the given environments. Briefly speaking, the robust version is to \textbf{a)} run the step I to III of ISR-Cov over $N \leq \binom E 2$ pairs of environments with distinct sample covariances, leading to $N$ $d_c$-dimensional subspaces obtained through Algorithm \ref{algo:isr-cov}, \textbf{b)} we find the stable $d_c$-dimensional subspace, and use it as the invariant-feature subspace that we train the following classifier. Specifically, we achieve b) by computing the flag-mean \citep{marrinan2014finding} over the set of $P'$ (defined in \eqref{eq:trans-matrix-isr-cov}) obtained from the $N$ selected pairs of environments. Compared with the original ISR-Cov, this robust version makes use of training data more efficiently (e.g., it uses more than one pair of training environments) and is more robust in the finite-data case. We implement this robust version of ISR-Cov in our experiments in Sec.~\ref{sec:exp}.

\setlength{\textfloatsep}{5pt}
\begin{algorithm}[t!]
\caption{ISR-Cov}\label{algo:isr-cov}
\begin{algorithmic}
\STATE {\bfseries Input:} Data of all training environments, $\{\mathcal D_e\}_{e\in [E]}$.
\FOR{$e = 1,2,\dots,E$}     
    \STATE Estimate the sample covriance of $\{x|(x,y)\in \mathcal D_e, y=1\}$ as $\Sigma_e \in \bR^{d \times d}$
\ENDFOR
\STATE \textbf{I.} Select a pair of environments $e_1,e_2$ such that $\Sigma_1 \neq \Sigma_2$, and compute their difference, $\Delta \Sigma \coloneqq \Sigma_{e_1} - \Sigma_{e_2}$
\STATE \textbf{II.} Eigen-decompose $\Delta\Sigma$ to obtain eigenvectors $\{P_1,...,P_d\}$  with eigenvalues $\{\lambda_1,...,\lambda_d\}$
\STATE \textbf{III.} Stack $d_c$ eigenvectors of eigenvalues with lowest absolute values to obtain a matrix $P'\in \bR^{d_c\times d}$
\STATE \textbf{IV.} Fit a linear classifier (with $w\in \bR^{d_c}$, $b\in \bR$) by ERM over all training data with transformation $x\mapsto P'x$
\STATE Obtain a predictor $f(x) =  w^\T P' x + b$
\end{algorithmic}
\end{algorithm}

\begin{figure}[t!]
\begin{center}
\centerline{\includegraphics[width=.5\columnwidth]{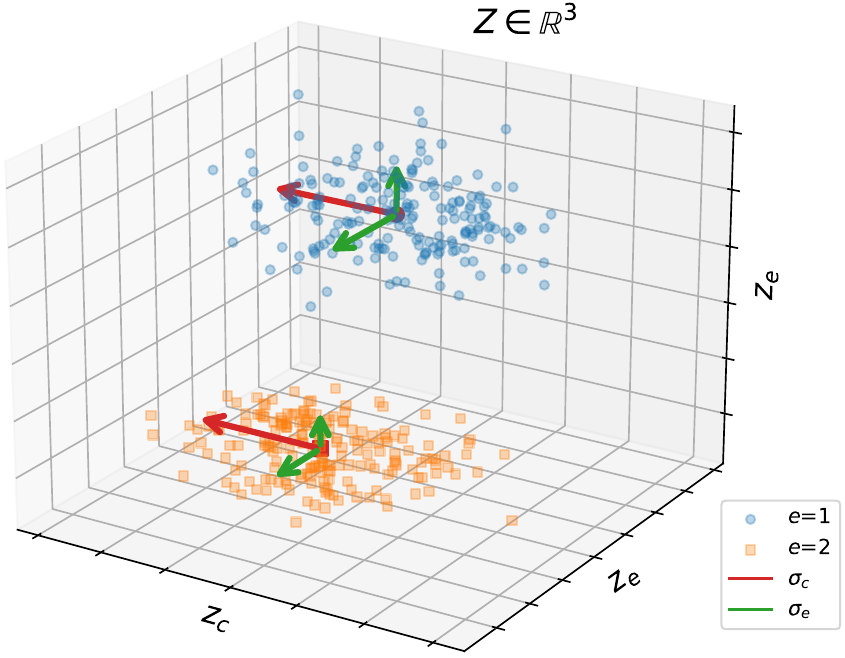}}
\vskip -0.1in
\caption{An example for ISR-Cov, where $d_c\mathrm{=}1$, $d_s\mathrm{=}2$, $E\mathrm{=}2$. In this latent feature space of $z\in\bR^3$, there is one dimension of $z_c$ and the rest two of $z_e$. 
}
\label{fig:ISR-cov-demo}
\end{center}
\vskip -0.15in
\end{figure}

\paragraph{Geometric Interpretation}
We provide an geometric interpretation of ISR-Cov with a 3D example in Fig.~\ref{fig:ISR-cov-demo}, where $d_c\mathrm{=}1$, $d_s\mathrm{=}2$, $E\mathrm{=}2$. For either environment $e\in\{1,2\}$, the covariance of its class-conditional latent-feature distribution, $\begin{bmatrix}
\sigma_c^2 I_{d_c} & 0\\
0 & \sigma_e^2 I_{d_s}
\end{bmatrix}$, is \textit{anisotropic}: the variance $\sigma_c$ along invariant-feature dimensions is constant, while $\sigma_e$ along the spurious-feature dimensions is various across $e\in \{1,2\}$ (ensured by Assumption \ref{assum:non-degenerate-cov}). Though the transformation $R$ is applied to latent features, ISR-Cov still can identify the subspace spanned by invariant-feature dimensions in the latent-feature space by utilizing this anisotropy property. 

\paragraph{$\cO(1)$ Environment Complexity}
In the infinite-sample setting, we prove below that as long as there are at least two training environments that satisfy Assumption \ref{assum:non-degenerate-cov} and \ref{assum:full-rank-transform}, ISR-Cov is guaranteed to learn the invariant optimal predictor. This is the minimal possible environment complexity, since spurious and invariant-features are indistinguishable with only one environment. Notably, unlike IRM, a) ISR-Cov does not require Assumption \ref{assum:non-degenerate-mean}, and b) ISR-Cov has a global convergence guarantee. The proof is in \cref{supp:proof:isr-cov}.

\begin{theorem}[ISR-Cov]\label{thm:isr-cov} Suppose $E\geq 2$ and the data size of each environment is infinite, i.e., $|\mathcal D_e| \mathrm{\rightarrow} \infty$ for $e\mathrm{=}1,\dots,E$. Eigen-decomposing $\Delta \Sigma$ defined in \eqref{eq:delta-sigma-expression}, the obtained eigenvectors $\{P_1,\dots, P_d\}$ with corresponding eigenvalues $\{\lambda_1,\dots, \lambda_d\}$ (ascendingly ordered by absolute values) satisfy
\begin{align*}
    \forall 1\leq i \leq d_c, ~\lambda_i = 0 \quad \text{and} \quad  \forall d_c < i \leq d, ~\lambda_i \neq 0
\end{align*}
The eigenvectors corresponding to these zero eigenvalues, i.e., $\{P_1,\dots, P_{d_c}\}$, can recover the subspace spanned by the invariant latent feature dimensions, i.e.,
\begin{align}\label{eq:thm:ISR-Cov:span-equality}
\mathrm{Span}(\{P_1^\T R,\dots, P_{d_c}^\T R\}) = \mathrm{Span}(\{\mathbf{\hat d_c^1},\dots, \mathbf{\hat d_c^{d_c}}\})
\end{align}
where $\mathbf{\hat d_c^i}$ is the unit-vector along the $i$-th coordinate in the latent feature space for $i=1,\dots, d_c$. Then, the classifier $f$ fitted with ERM to training data transformed by $x\mapsto [P_1,\dots,P_{d_c}]^\T x$ is guaranteed be the invariant optimal predictor, i.e., $f = h^*$, where $h^*$ is defined in \eqref{eq:optimal-inv-pred}.
\end{theorem}

\subsection{ISR-Multiclass} \label{sec:isr-multiclass}

Now, we introduce invariant-feature subspace recovery for multi-class classification. As discussed in section \ref{sec:setup-multiclass},  the binary \textit{binary} target variable induces \textit{symmetric} means for both the invariant and spurious features. This is not necessary, and we now consider a more general scenario. Moreover, in the more realistic multi-class setting, it is unclear whether there exists any relation between the number of classes and the number of training environments required to recover invariant features. Further, \cite{singla2021salient} demonstrates that features that are spurious for a given target label may be the core features for another. One could apply class-specific transformations to mitigate this. However, this is infeasible as we do not know the class labels during testing. Thus, we would now like to answer the following question:
\begin{center}
    \textit{In the context of linear causal models, can the number of classes compensate for the number of training environments in recovering the invariant features?}
\end{center}

In order to answer the above question, we propose \textbf{ISR-Multiclass}, an extension of the ISR-Mean algorithm that provably recovers the invariant subspace in $\lceil d_s/k \rceil + 1$ environments, where $d_s$ is the number of spurious features (i.e., the dimensionality of the spurious-feature subspace) and $k$ is the number of classes. Note that our result improves over the original environmental complexity of $d_s+1$ for the binary classification problem\footnote{When $k=2$, the environmental complexity in ISR-Mean is $d_s+1$ instead of $d_s / 2 +1$ because of one specific assumption on the symmetry of the conditional feature distributions, which we also remove in this section.}, hence it provides an affirmative answer to the above problem. Furthermore, our result shows that the benefits of $k$-class classification problems help to reduce the environmental complexity by an order of $k$.

The ISR-Multiclass algorithm is outlined in Algorithm \ref{algo:isr-multiclass}. First, the detailed steps for the algorithm are presented, followed by a theorem formally stating the improvement in the environment complexity.

\setlength{\textfloatsep}{10pt}
\begin{algorithm}[htb]
\caption{ISR-Multiclass}\label{algo:isr-multiclass}
\begin{algorithmic}
\STATE {\bfseries Input:} Data of all training environments, $\{\mathcal D_e\}_{e\in [E]}$ across all classes $y \in \{y_1, y_2, \dots , y_k\}$.
\FOR{$y = y_1,y_2,\dots,y_k$}     
\FOR{$e = 1,2,\dots,E$}     
    \STATE Estimate the sample mean of $\{x|(x,y)\in \mathcal D_e, y=y_k\}$ as $\bar x_{ke}\in \bR^{d}$
\ENDFOR
\ENDFOR
\STATE \textbf{I.} Construct $k$ matrices $\mathcal M_k \in \bR^{E \times d}$ with the $e$-th row of the $k$-th matrix as $\bar{x}_{ke}^\T$ for $e\in [E]$
\STATE \textbf{II.} Apply PCA to each $\mathcal M_k$ to obtain a set of eigenvectors. Choose eigenvectors corresponding to $E - 1$ highest eigenvalues to construct $P_k 
\mathrm{\coloneqq} \begin{bmatrix}
    P_{1k} | P_{2k} | \cdots | P_{(E-1)k}
    \end{bmatrix} \in \mathbb{R}^{d \times (E-1)}$
\STATE \textbf{III.} Stack each $P_k$ to obtain $\mathcal M_{total} \mathrm{\coloneqq}$ $\begin{bmatrix}
    P_{1} | P_{2} | \cdots | P_{k}
    \end{bmatrix} \in \mathbb{R}^{d \times (E-1)k}$
\STATE \textbf{IV.} Apply SVD of $\mathcal{M}_{total}$ to 
obtain singular vectors $\{P_1^{\prime},...,P_d^{\prime}\}$ with singular values $\{\lambda_1,...,\lambda_d\}$
\STATE \textbf{V.} Stack $d_s$ singular vectors with the highest singular values to obtain a transformation matrix $P'\in \bR^{d_s\times d}$ 
\STATE \textbf{VI.} Apply transformation $x\mapsto P''x$ on the training data and fit a linear classifier (with $w\in \bR^{d_c}$, $b\in \bR$)
\STATE Obtain a predictor $f(x) =  w^\T P'' x + b$
\end{algorithmic}
\end{algorithm}

\paragraph{I. Estimating Sample Means for Every Environment and Every Class}

Construct the following matrix $\mathcal{M}_k$ for class $k$ where each row contains the sample mean conditioned on a given environment $e$, for class $k$. In other words, each row is $\bar{x}_{ek} = A \mu_{k} + B \mu_{ke}$. Note that in the infinite sample setting considered in this work, this is exactly the mean as per Equation \eqref{eq:transformed-gaussian}. 
\begin{align}\label{eq:def-Mk}
    \mathcal{M}_k \mathrm{\coloneqq} \begin{bmatrix}
    \bar{x}_{1k}^\T\\
    \vdots\\
    \bar{x}_{Ek}^\T
    \end{bmatrix} 
\end{align}

\paragraph{II. Conduct PCA for every $\mathcal{M}_k$} 

For every class, conduct PCA to obtain eigenvectors
$\{P_i\}_{i=1}^d$ corresponding to eigenvalues $\{\lambda_i\}_{i=1}^d$. By assumption \ref{assum:non-degenerate-mean-multiclass}, this step yields $E-1$ eigenvectors (-1 from the mean centering in PCA) which correspond to non-zero eigenvalues. 
\begin{align}\label{eq:def-Pk}
    P_k \mathrm{\coloneqq} \begin{bmatrix}
    P_{1k} | P_{2k} | \cdots | P_{(E-1)k}
    \end{bmatrix} \in \mathbb{R}^{d \times (E-1)}
\end{align} 
$P_{ik}$ is the $i^{th} $eigenvector corresponding to a non-zero eigenvalue in the phase transition of $\mathcal{M}_k$. Thus, $P_k$ recovers spurious dimensions corresponding to class $k$, as follows from ISR-Mean. Note that unlike ISR-Mean, the condition $E > d_s$ \textit{is not imposed} as information from a single class may not be sufficient to recover all spurious (and thus invariant) features. 

\paragraph{III. Stack all $P_k$ and take SVD (Singular Value Decomposition)}
After obtaining $P_k$ for every class $k$, stack all $P_k$ to obtain a new matrix $\mathcal{M}_{total}$ as follows:
\begin{align}\label{eq:def-Mtotal}
    \mathcal{M}_{total} \mathrm{\coloneqq} \begin{bmatrix}
    P_{1} | P_{2} | \cdots | P_{k}
    \end{bmatrix} \in \mathbb{R}^{d \times (E-1)k}
\end{align}

Next, take the SVD of $P_k$. Note that this step is motivated by the 
flag-mean~\citep{marrinan2014finding} to find the \textit{common spurious subspace} for all class labels.

\paragraph{IV. Extract Spurious Feature Subspace}
In \Cref{thm:isr-multiclass}, we prove that SVD of $\mathcal{M}_{total}$ leads to $d_s$ non zero singular values, where the corresponding vectors $\{P_1^{\prime},\dots,P_{d_s}^{\prime}\}$ recover the underlying spurious subspace. Stack these vectors as a matrix $P'$:
\begin{align}\label{eq:trans-matrix-spu-isr-multiclass}
    P' \coloneqq [P_1^{\prime},\dots, P_{d_s}^{\prime}]^\T \in \bR^{d_s \times d}
\end{align}

\paragraph{V. Train a Classifier in the Null Space of Spurious-Feature Subspace} 
The final step involves training a classifier in the \textit{null space} of the extracted spurious feature subspace, which is the invariant-feature subspace:
\begin{align}\label{eq:trans-matrix-inv-isr-multiclass}\mathrm{Col}(P'') = NullSpace(P') \in \bR^{d \times d_c} 
\end{align}

\begin{theorem}[ISR-Multiclass]\label{thm:isr-multiclass}
Suppose $E \geq \lceil d_s/k\rceil + 1$ and the data size of each environment is infinite, i.e., $|\mathcal D_e| \mathrm{\rightarrow} \infty$ for $e\mathrm{=}1,\dots,E$. On performing \textbf{SVD} for $\mathcal{M}_{total}$ (defined in \eqref{eq:def-Mtotal}), let $\{\lambda_1,\dots, \lambda_d\}$ denote the set of singular values obtained in descending order. Then, the top $d_s$ singular values are strictly positive:
\begin{align*}
    \forall 1\leq i \leq d_s, ~\lambda_i > 0 
\end{align*}
The singular vectors corresponding to the top $d_s$ strictly positive singular values $P_{positive} =  \{P_1^{\prime},\dots, P_{d_s}^{\prime}\}$ correspond to the $d_s$ spurious dimensions:
\begin{align}
    \text{Span}(\{P_i^{'\T} R: P_i^{'} \in P_{positive}\}) = \mathrm{Span}(\{\mathbf{\hat d_s^1},\dots, \mathbf{\hat d_s^{d_s}}\})
\end{align}
where the $\mathbf{\hat d_s^i}$ denotes the unit vector along the $i^{th}$ dimension of the latent spurious feature space i.e.\ $i = 1, 2, \dots, d_s$. 
Consequently, the $\text{NullSpace}([P_1^{\prime},\dots,P_{d_s}^{\prime}]^\T \in \mathbb{R}^{d_s \times d}) = \text{Col}([P_1^{\prime\prime},\dots,P_{d_c}^{\prime\prime}]^\T) \in \mathbb{R}^{d_c \times d} $ corresponds to a transformation matrix which recovers the invariant-feature subspace. Then, the classifier $f$ fitted with ERM to training data transformed by $x\mapsto [P_1^{\prime\prime},\dots,P_{d_c}^{\prime\prime}]^\T x$ is guaranteed to be the invariant optimal predictor, i.e., $f = h^*$, where $h^*$ is defined in \eqref{eq:optimal-inv-pred}.
\end{theorem} 

\paragraph{Global Convergence} 
The ISR-Multiclass algorithm involves applying an additional SVD operation over ISR-Mean, which makes the entire process optimization-free: a classifier trained on these features is globally optimal.

\paragraph{Environment Complexity} The environment complexity of ISR-Multiclass is $\lceil d_s/k\rceil + 1$ (detailed proof in \ref{supp:proof:isr-multiclass}). The key observation here is that the column space of each $P_i$ matrix for $i\in[k]$ only consists of the span of $\{\mu_{ie}\}_{e=1}^E$. In order to identify the subspace of spurious features, one needs at least $d_s$ linearly independent components. Hence, the only need is to ensure that $(E-1)k\geq d_s$ so that the column space of $\mathcal{M}_{total}$ is full rank. Solving this inequality leads us to the desired bound on the environment complexity. This implies that one can leverage information from \textit{multiple classes} to \textit{reduce} the environment complexity by a factor of $1/k$, as compared to $d_s+1$ proposed in \Cref{sec:isr-mean}, while only relying on the $1^{st}$ order moments of the data generating distribution. Intuitively, ISR-Multiclass leverages \textit{additional} information from multiple classes to find the common spurious feature subspace in lesser number of environments.

\subsection{Regression}\label{sec:isr-regression}
We now study the problem of invariant-feature subspace recovery in the setting of regression. In regression, the target is continuous, which makes it non-trivial to extend the current framework for provable recovery of the invariant-feature subspace in the absence of a discrete class label. Thus, through this section, we answer the following question:
\begin{center}
    \textit{In the context of linear causal models, can we provably recover the invariant-feature subspace when the target is continuous?}
\end{center}

To this end, we first identify a \textit{new} causal model which is more natural for studying regression problems and then propose ISR-Regression i.e.\ \textbf{I}nvariant Feature \textbf{S}ubspace \textbf{R}ecovery for \textbf{Regression}. ISR-Regression extends the notion of invariant-feature subspace recovery under the new causal model for regression and requires $d_s + 1$ environments to provably recover the invariant-feature subspace.

\paragraph{Note on causal model for regression} 
Our proposed causal model for studying regression problems is outlined in~\Cref{fig:cgraph-regression}. This causal model specifies that the 
domain-invariant features $z_c$ causally determine $y$, which is \textit{natural} for regression problems: one typically observes a dependent variable $y$ corresponding to a set of features $x$. This is in contrast to previous works modeling regression in the anti-causal setting~\citep{ahuja2021invariance} where the causal mechanism between the features and the labels are reversed. Next, there may be spurious (environment-dependent) features $z_e$ which might be correlated with the target $y$ in some environments but are independent of $y$ given $z_c$, i.e.\ $y \perp z_e | z_c$. Thus, the aim is to recover the 
domain invariant feature subspace corresponding to $z_c$ in order to generalize well to new environments during testing, where the correlation with $z_e$ may change or may not even exist.

\subsubsection{ISR-Regression}
This section introduces the ISR-Regression algorithm, outlined in Algorithm~\ref{algo:isr-regression}. We first discuss the detailed steps for applying ISR-Regression and then present a theorem to provide a formal guarantee on environment complexity.

\setlength{\textfloatsep}{10pt}
\begin{algorithm}[tb]
\caption{ISR-Regression}\label{algo:isr-regression}
\begin{algorithmic}
\STATE {\bfseries Input:} Data of all training environments, $\{\mathcal D_e\}_{e\in [E]}$.
\FOR{$e = 1,2,\dots,E$}    
    \STATE 
    \STATE Estimate the sample mean of $\{x|(x,y)\in \mathcal D_e\}$ as $\bar x_e\in \bR^{d}$
\ENDFOR
\STATE \textbf{I.} Construct matrix $\mathcal M \in \bR^{E \times d}$ with the $e$-th row as $\bar{x}_e^\T$ for $e\in [E]$
\STATE \textbf{II.} Apply PCA to $\mathcal M$ to obtain eigenvectors $\{P_1,...,P_d\}$ with eigenvalues $\{\lambda_1,...,\lambda_d\}$
\STATE \textbf{III.} Choose $d_s$ eigenvectors corresponding to the highest eigenvalues to and stack them to obtain a transformation matrix $P''\in \bR^{d_s\times d}$. Take the null space of $P''$ to obtain transformation matrix $P'\in \bR^{d_c\times d}$
\STATE \textbf{IV.} Fit a linear classifier (with $w\in \bR^{d_c}$, $b\in \bR$) by ERM over all training data with transformation $x\mapsto P'x$
\STATE Obtain a predictor $f(x) =  w^\T P' x + b$
\end{algorithmic}
\end{algorithm}

\paragraph{Intuition for Algorithm \ref{algo:isr-regression}} 
Intuitively, Algorithm \ref{algo:isr-regression} can be viewed as applying ISR-Mean \ref{algo:isr-mean} over the \textit{entire} dataset as opposed to conditioning on a single class. 
Further, the current data model in \ref{sec:setup-regression} for regression considers $y$ to be the dependent variable (as is typically assumed in regression). Conducting the PCA on the global input means still helps us in the provable recovery of the spurious and thus invariant-feature subspace. Geometrically, instead of having a particular kind of data distribution \textit{per class} like in ISR-Mean, we now have this structure globally.

Next, let us look at the detailed steps for conducting ISR-Regression:

\paragraph{I. Estimate Sample Means per Environment} First, construct the matrix $\cM \in \bR^{E \times d}$ where every row of the matrix is the mean $\bar{x}_e$ i.e.\ mean of samples belonging to a specific environment $e$. Based on the sampling of $x$ as per \eqref{eq:def-linear-transform-regression}
this row represents an estimate of the distribution mean $A\mu_c + B\mu_e$ where:
\begin{align}
    \mu_e = W_{cs}^e\mu_c + b_e
\end{align}
Now, we know that $\bar x_e$ satisfies $\bar{x}_e = A \mu_c + B \mu_e$. Combining this with the fact that $R = [A ~ B]$, the matrix $\mathcal M$ can be expressed as
\begin{align}\label{eq:def-M-regression}
    \mathcal M \mathrm{\coloneqq} \begin{bmatrix}
    \bar{x}_1^\T\\
    \vdots\\
    \bar{x}_E^\T
    \end{bmatrix} \mathrm{=} \begin{bmatrix}
    \mu_c^\T A^\T \mathrm{+} \mu_1^\T B^\T\\
    \vdots\\
    \mu_c^\T A^\T \mathrm{+} \mu_E^\T B^\T\end{bmatrix}
    \mathrm{=}{\overbrace{\begin{bmatrix}
    \mu_c^\T ~~\mu_1^\T\\
    \vdots~~~~~\vdots\\
    \mu_c^\T ~~ \mu_E^\T
    \end{bmatrix}}^{\mathcal U^\T \coloneqq}} R^\T
\end{align}

\paragraph{II. Apply PCA on $\cM$} After computing $\cM$ as defined above, apply PCA on $\cM$. This involves mean centering $\cM $ to obtain  $\wt{\cM}$ where the sample mean is subtracted from every row of $\cM$. This is followed by eigen-decomposition of the sample covariance matrix of $\wt{\cM}$ which is $\wh \Sigma_\cM \coloneqq \frac{1}{E} \wt{\cM}^\T \wt{\cM}$. This yields a set of $d$ eigenvectors $\{P_1, P_2, \cdots , P_d\}$ and corresponding eigenvalues $\{\lambda_1, \lambda_2, \cdots , \lambda_d\}$.

\paragraph{III. Obtain Invariant-Feature Subspace from Eigenvectors} Stack the eigenvectors corresponding to the highest $d_s$ eigenvalues to obtain a transformation matrix $P'' \in \bR^{d_s \times d}$. Then, take the null space of this matrix to obtain $P' \in \bR^{d_c \times d}$. 

Note that when $E \geq d_s + 1$, one can choose the \textit{lowest} $d_c$ eigenvalues and stack the corresponding eigenvectors to obtain a transformation directly into the invariant-feature subspace. However, when $E < d_s + 1$, this will not be equivalent to the presented algorithm, a discussion of which is discussed in section \ref{supp:proof:isr-mean}.

\paragraph{IV. Train a Classifier in the Invariant-Feature Subspace} Once the transformation matrix $P'$ has been computed, transform the input training data $x \mapsto P'x$ and fit a linear regressor on top of this transformed data to obtain a predictor $f(x) = w_c^\top x' + b$.

\begin{theorem}[ISR-Regression]\label{thm:isr-regression}
Suppose $E > d_s$ and the data size of each environment is infinite, i.e., $|\mathcal D_e| \mathrm{\rightarrow} \infty$ for $e\mathrm{=}1,\dots,E$. On performing \textbf{PCA} for $\cM$ (defined in \eqref{eq:def-M-regression}), let the  $\{\lambda_1,, \lambda_2, \cdots, \lambda_d\}$ denote eigenvalues sorted in ascending order and $\{P_1, P_2, \cdots, P_d\}$ denote the corresponding eigenvectors. Then, the following property holds:
\begin{align}
     \lambda_i = \begin{cases}
      0  &~ \text{if}~ 1 \leq i \leq d_c \\    
      > 0 &~\text{if}~ d_c < i \leq d
    \end{cases}
\end{align}
The eigenvectors $P_{zero}$ corresponding to the $d_c$ zero eigenvalues respectively can recover the invariant-feature subspace:
\begin{align}
    \text{Span}(\{P_i^\T R: P_i \in P_{zero}\}) = \mathrm{Span}(\{\mathbf{\hat d_c^1},\dots, \mathbf{\hat d_c^{d_c}}\})
\end{align}
where the $\mathbf{\hat d_c^i}$ denotes the unit vector along the $i^{th}$ dimension of the latent invariant feature space i.e.\ $i = 1, 2, \dots, d_c$. A regressor $f$ fitted with ERM to training data transformed by $x\mapsto [P_1,\dots,P_{d_c}]^\T x$ is guaranteed to be the invariant optimal predictor, i.e., $f = h^*$, where $h^*$ is defined in \eqref{eq:optimal-inv-pred}.
\end{theorem}

The proof can be found in Appendix \ref{supp:proof:isr-regression}. 

\paragraph{Optimality} The ISR-Regression algorithm is globally optimal: applying PCA does not involve local search and other steps in the algorithm are optimization free. Once the invariant-feature subspace is extracted, performing linear regression recovers the globally optimal predictor due to the convex loss function.

\paragraph{Environment Complexity} The environment complexity of ISR-Regression is $d_s + 1$, as will be proved in section \ref{supp:proof:isr-regression}. This matches that of the original ISR-Mean algorithm in the binary classification setting.

\section{Experiments}\label{sec:exp}
We conduct experiments on both synthetic and real datasets to examine our proposed algorithms. 

\subsection{Binary Classification}\label{sec:exp-bin-classification}

\subsubsection{Synthetic Datasets: Linear Unit-Tests}  \label{sec:lut}
We adopt a set of synthetic domain generalization benchmarks, Linear Unit-Tests \citep{aubin2021linear}, which is proposed by authors of IRM and is used in multiple recent works \citep{koyama2020out,khezeli2021invariance,du2021beyond}.
Specifically, we take four classification benchmarks from the Linear Unit-Tests, which are named by \citet{aubin2021linear} as Example-2/2s/3/3s. Example-2 and 3 are two binary classification tasks of Gaussian linear data generated in processes similar to the setup of Sec. \ref{sec:setup}, and they have identity transformation, $R=I$ (see the definition of $R$ in \eqref{eq:def-linear-transform}), while Example 2s/3s are their counterparts with $R$ as a random transformation matrix. However, Example-3/3s do not satisfy Assumption \ref{assum:non-degenerate-cov}, thus cannot properly examine our ISR-Cov. Hence, we construct variants of Example-3/3s satisfying Assumption \ref{assum:non-degenerate-cov}, which we name as Example-3'/3s', respectively. We provide specific details of these benchmarks in \cref{supp:exp:synthetic-setup}.

\textit{Example-2}: The data generation process for Example-2 follows the Structual Causal Model \citep{peters2015causal}, where $P(Y|\mu_c)$ is invariant across environments.

\textit{Example-3}: It is similar to our Gaussian setup in \ref{sec:setup}, except that $\sigma_e\equiv \sigma_c=0.1$, breaking Assumption \ref{assum:non-degenerate-cov}. In this example, $P(\mu_c|Y)$ is invariant across environments.

\textit{Example-3'}: We modify Example-3 slightly such that $\sigma_c=0.1$ and $\sigma_e \sim \mathrm{Unif}(0.1, 0.3)$. All the rest settings are identical to Example-3.

\textit{Example-2s/3s/3s'}: A random orthonormal projection matrix $R = [A,~B] \in \bR^{d\times d}$ (see the definition in \eqref{eq:def-linear-transform}) is applied to the original Example-2/3/3' to scramble the invariant and spurious latent feature, leading to Example-2s/3s/3s' with observed data in the form of $x = Az_c + B z_e$.
\begin{figure*}[t!]
\begin{center}
\centerline{\includegraphics[width=1\linewidth]{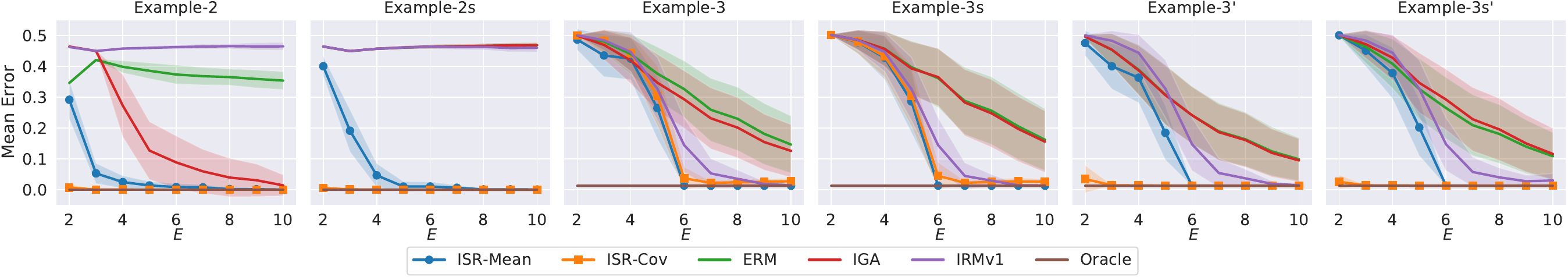}}
\caption{Test results on Linear Unit-Tests (first 4 plots) and its variants (last 2 plots), where $d_c = 5, d_s=5$, and $E=2,\dots, 10$.
}
\label{fig:linear-unit-tests}
\end{center}
\end{figure*}

\paragraph{Implementation}
For baseline algorithms, we directly adopt their implementations by \citet{aubin2021linear}. We implement ISRs following \cref{algo:isr-mean} and \ref{algo:isr-cov}, where the last step of fitting predictors is done by the ERM implementation of \citet{aubin2021linear}, which optimizes the logistic loss with an Adam optimizer \citep{adam}. More details are provided in \cref{supp:exp}.

\paragraph{Evaluation Procedures}~
Following \citet{aubin2021linear}, we fix $d\mathrm{=}10$, $d_c \mathrm{=} 5$, $d_s\mathrm{=}5$, and increase $E$, the number of training environments, from 2 to 10, with 10K observed samples per environment. Each algorithm trains a linear predictor on these training data, and the predictor is evaluated in $E$ test environments, each with 10K data. The test environments are generated analogously to the training ones, while the spurious features $z_e$ are randomly shuffled across examples within each environment. The mean classification error of the trained predictor over $E$ test environments is evaluated.

\paragraph{Empirical Comparisons}~
We compare our ISRs with several algorithms implemented in \citet{aubin2021linear}, including IRMv1, IGA (an IRM variant by \citet{koyama2020out}), ERM and Oracle (the optimal invariant predictor) on the datasets. We repeat the experiments over 50 random seeds and plot the mean errors of the algorithms. Fig. \ref{fig:linear-unit-tests} shows the results of our experiment on these benchmarks: 
\textbf{a)} On Example-2/2s, our ISRs reach the oracle performance with a small $E$ (number of training environments), significantly outperforming other algorithms.
\textbf{b)} On Example-3/3s, ISRs reach the oracle performance as $E > 5=d_s$, while IRM or others need more environments to match the oracle. 
\textbf{c)} On Example-3'/3s', ISR-Cov matches the oracle as $E \geq 2$, while the performance of all other algorithms does not differ much from that of Example-3/3s.

\paragraph{Conclusions}~ Observing these results, we can conclude that:
\textbf{a)} ISR-Mean can stably match the oracle as $E > d_s$, validating the environment complexity proved in \cref{thm:isr-mean}. 
\textbf{b)} ISR-Cov matches the oracle as $E\geq 2$ in datasets satisfying Assumption \ref{assum:non-degenerate-cov} (i.e., Example-2/2s/3'/3s'), corroborating its environment complexity proved in \cref{thm:isr-cov}.

\begin{figure*}[ht!]
\begin{center}
\centerline{\includegraphics[width=1\linewidth]{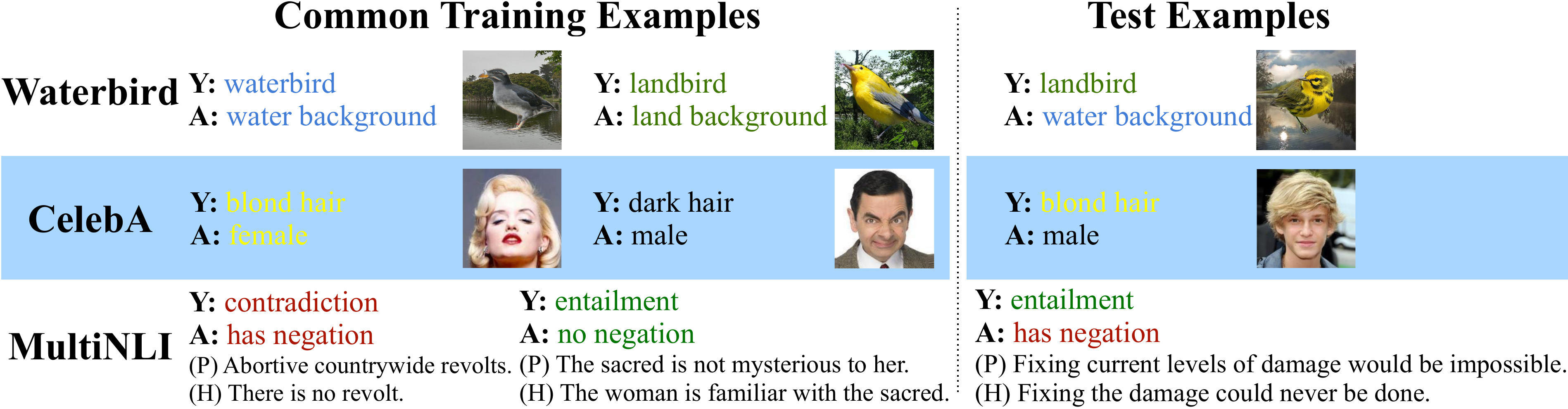}}
\vskip -0.1in
\caption{Representative examples of the three real datasets we use. The spurious correlation between the label (\textbf{Y}) and the attribute (\textbf{A}) in the training data does not hold in the test data.
}
\label{fig:real-datasets}
\end{center}
\vspace{-1em}
\end{figure*}

\vspace{.5em}
\subsubsection{Real Datasets} 

\begin{table*}[ht!]
\centering
    \resizebox{.99\textwidth}{!}{%
    \centering  
    \setlength{\tabcolsep}{4.5pt}
\begin{tabular}{@{}cclccccccc@{}}
\toprule
\multicolumn{1}{c}{\textbf{Dataset}}    & \multicolumn{1}{c}{\textbf{Backbone}}&
\multicolumn{1}{c}{\textbf{Algorithm}} &
\multicolumn{3}{c}{\textbf{Average Accuracy}} & 
\multicolumn{3}{c}{\textbf{Worst-Group Accuracy}} \\ \cmidrule(l){4-6} \cmidrule(l){7-9} 
& \multicolumn{2}{c}{\text{~}}&\textbf{ Original}& \textbf{ ISR-Mean}& \textbf{ ISR-Cov}& \textbf{ Original}& \textbf{ ISR-Mean} &\textbf{ ISR-Cov}\\ 
\midrule
\multicolumn{1}{c}{Waterbirds} 
& ResNet-50 &ERM& 86.66$\pm$0.67& 87.87$\pm$0.80& \textbf{90.47$\pm$0.33} ~&~
62.93$\pm$5.37& 76.10$\pm$1.11& \textbf{82.46$\pm$0.55}\\
& & Reweighting & 91.49$\pm$0.46& \textbf{91.77$\pm$0.52}& 91.63$\pm$0.44 ~&~
87.69$\pm$0.53& 88.02$\pm$0.42& \textbf{88.67$\pm$0.55}\\
& & GroupDRO& 92.01$\pm$0.33& 91.74$\pm$0.35& \textbf{92.25$\pm$0.27} ~&~
90.79$\pm$0.47& 90.42$\pm$0.61& \textbf{91.00$\pm$0.45}\\ 
\midrule
\multicolumn{1}{c}{CelebA}
&ResNet-50& ERM& \textbf{95.12$\pm$0.34}& 94.34$\pm$0.12 &90.12$\pm$2.59 ~&~
46.39$\pm$2.42& 55.39$\pm$6.13& \textbf{79.73$\pm$5.00}\\
&& Reweighting & \textbf{91.45$\pm$0.50}& 91.38$\pm$0.51 & 91.24$\pm$0.35 ~&~
84.44$\pm$1.66& \textbf{90.08$\pm$0.50}& 88.84$\pm$0.57\\
&& GroupDRO&\textbf{91.82$\pm$0.27}& 91.82$\pm$0.27& 91.20$\pm$0.23 ~&~
88.22$\pm$1.67& \textbf{90.95$\pm$0.32}& 90.38$\pm$0.42\\ 
\midrule
\multicolumn{1}{c}{MultiNLI}
&BERT& ERM&\textbf{82.48$\pm$0.40}& 82.11$\pm$0.18& 81.28$\pm$0.52 ~&~
65.95$\pm$1.65& 72.60$\pm$1.09& \textbf{74.21$\pm$2.55} \\
&& Reweighting & \textbf{80.82$\pm$0.79} & 80.53$\pm$0.88& 80.73$\pm$0.90 ~&~ 
64.73$\pm$0.32& \textbf{67.87$\pm$0.21}& 66.34$\pm$2.46\\
&& GroupDRO&\textbf{81.30$\pm$0.23}& 81.21$\pm$0.24& 81.20$\pm$0.24 ~&~
78.43$\pm$0.87& \textbf{78.95$\pm$0.95}& 78.91$\pm$0.75\\ \bottomrule
\end{tabular}}
\caption{Test accuracy(\%) with standard deviation of ERM, Re-weighting and GroupDRO over three datasets. We compare the accuracy of original trained classifiers vs. ISR-Mean post-processed classifiers. The average accuracy and the worst-group accuracy are both presented. Bold values mark the higher accuracy over Original vs. ISR-Mean for a given algorithm (e.g., ERM) and a specific metric (e.g., Average Acc.).} 
    \label{tab:real-datasets}

\end{table*}

 We adopt three datasets that \citet{sagawa2019distributionally} proposes to study the robustness of models against spurious correlations and group shifts. See Fig. \ref{fig:real-datasets} for a demo of these datasets. Each dataset has multiple spurious attributes, and we treat each spurious attribute as a single environment.

\textit{Waterbirds} \citep{sagawa2019distributionally}: This is a image dataset built from the CUB \citep{wah2011caltech} and Places \citep{zhou2017places} datasets. The task of this dataset is the classification of waterbirds vs. landbirds. Each image is labelled with class $y\in \mathcal Y = \{\textit{waterbird, landbird}\}$ and environment $e\in \mathcal E= \{\textit{water background, land background}\}$. \citet{sagawa2019distributionally} defines 4 groups\footnote{Notice that the definition of environment in this paper is different from the definition of group in \citet{sagawa2019distributionally}.} by $\mathcal G = \mathcal Y \times \mathcal E$. There are 4795 training samples, and smallest group (waterbirds on land) only has 56.

\textit{CelebA} \cite{liu2015faceattributes}: This is a celebrity face dataset of 162K training samples. \citet{sagawa2019distributionally} considers a hair color classification task ($\mathcal Y = \{\textit{blond, dark}\}$) with binary genders as spurious attributes (i.e., $\mathcal E = \{\textit{male, female}\}$). Four groups are defined by $\mathcal G = \mathcal Y \times \mathcal E$, where the smallest group (blond-haired males) has only 1387 samples.

\textit{MultiNLI} \cite{williams2017broad}: This is a text dataset for natural language inference. Each sample includes two sentences, a hypothesis and a premise. The task is to identify if the hypothesis is contradictory to, entailed by, or neutral with the premise ($\mathcal Y = \{\textit{contradiction,neutral,entailment}\}$). \citet{gururangan2018annotation} observes a spurious correlation between $y\mathrm{=}\textit{contradiction}$ and negation words such as nobody, no, never, and nothing. Thus $\mathcal E \mathrm{=} \{\textit{no negation, negation}\}$ are spurious attributes (also environments), and 6 groups are defined by $\mathcal G \mathrm{=} \mathcal Y \mathrm{\times} \mathcal E$. There are 20K training data, while the smallest group (entailment with negations) has only 1521.

\paragraph{Implementation}~ We take three algorithms implemented by \citet{sagawa2019distributionally}: ERM, Reweighting, and GroupDRO. First, for each dataset, we train neural nets with these algorithms using the code and optimal hyper-parameters provided by \citet{sagawa2019distributionally} implementation, and early stop models at the epoch with the highest worst-group validation accuracy. Then, we use the hidden-layers of the trained models to extract features of training data, and fit ISR-Mean/Cov to the extracted features. Finally, we replace the original last linear layer with the linear classifier provided by ISR-Mean/Cov, and evaluate it in the test set. More details are provided in \cref{supp:exp}.

\paragraph{Empirical Comparisons}~
We compare trained models with the original classifier vs. ISR-Mean/Cov post-processed classifiers over three datasets. Each experiment is repeated over 10 random seeds. From the results in \cref{tab:real-datasets}, we can observe that: \textbf{a)} ISRs can improve the worst-group accuracy of trained models across all dataset-algorithm choices. \textbf{b)} Meanwhile, the average accuracy of ISR-Mean/Cov is maintained around the same level as the original classifier.

\begin{table*}[t]
\centering
    \resizebox{.99\textwidth}{!}{%
    \centering  
    \setlength{\tabcolsep}{4.5pt}
\begin{tabular}{@{}cclccccccc@{}}
\toprule
\multicolumn{1}{c}{\textbf{Dataset}}    & \multicolumn{1}{c}{\textbf{Backbone}}&
\multicolumn{1}{c}{\textbf{Algorithm}} &
\multicolumn{3}{c}{\textbf{Average Accuracy}} & 
\multicolumn{3}{c}{\textbf{Worst-Group Accuracy}} \\ \cmidrule(l){4-6} \cmidrule(l){7-9} 
& \multicolumn{2}{c}{\text{~}}&\textbf{\small Linear Probing}& \textbf{\small ISR-Mean}& \textbf{\small ISR-Cov}& \textbf{\small Linear Probing}& \textbf{\small  ISR-Mean} &\textbf{\small ISR-Cov}\\ 
\midrule
\multicolumn{1}{c}{Waterbirds} 
& CLIP (ViT-B/32) &ERM& 76.42$\pm$0.00& \textbf{90.27$\pm$0.09}& 76.80$\pm$0.01 ~&~
52.96$\pm$0.00& \textbf{71.75$\pm$0.39}& 55.76$\pm$0.00\\
&&Reweighting & 87.38$\pm$0.09& \textbf{88.23$\pm$0.12}& 88.07$\pm$0.05 ~&~
82.51$\pm$0.27& \textbf{85.13$\pm$0.22}& 83.33$\pm$0.00\\
\bottomrule
\end{tabular}}
\caption{Evaluation with CLIP-pretrained vision transformers. We compare ISR-Mean/ISR-Cov vs. linear probing in the Waterbird dataset, and report the test accuracy (\%) with standard deviation.} 
    \label{tab:CLIP}

\end{table*}

\vspace{.3em}
\subsubsection{Reduced Requirement of Environment Labels} 
\begin{figure}[t!]
\centering
\vspace{-.5em}
\includegraphics[width =0.5\columnwidth]{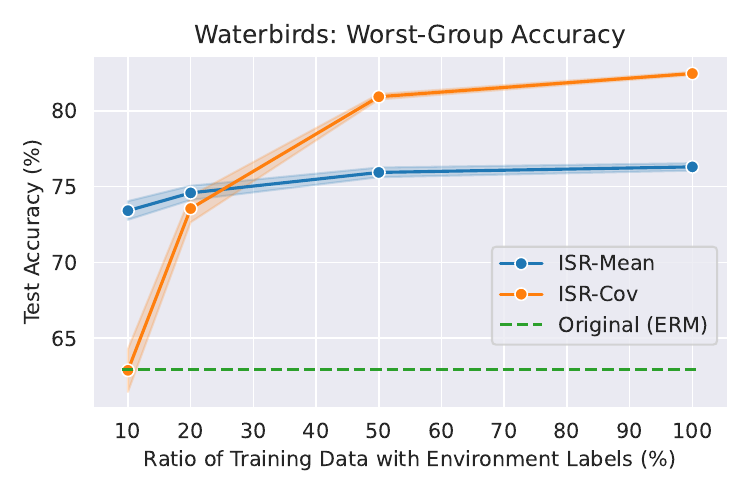}
\vspace{-1em}
\caption{Applying ISR-Mean/ISR-Cov to ERM-trained models with partially available environment labels in the Waterbirds dataset. The shading area indicates the 95\% confidence interval for mean accuracy.
}
\label{fig:partial-env}
\end{figure}

Algorithms such as GroupDRO are successful, but they require each training sample to be presented in the form $(x,y,e)$, where the environment label $e$ is usually not available in many real-world datasets. Recent works such as \citet{liu2021just} try to relieve this requirement. To this end, we conduct another experiment on Waterbirds to show that ISRs can be used in cases where only a part of training samples are provided with environment labels. Adopting the same hyperparameter as that of Table \ref{tab:real-datasets}, we reduce the available environment labels from $100\%$ to $10\%$ (randomly sampled), and apply ISR-Mean/Cov on top of ERM-trained models with the limited environment labels. We repeat the experiment over 10 runs for each of 10 ERM-trained models, and plot the mean accuracy in Fig. \ref{fig:partial-env}. We can observe that \textbf{a)} even with only $10\%$ environment labels, the worst-group accuracy of ISR-Mean attains $73.4\%$, outperforming the original ERM-trained classifier by a large margin of $10.5\%$, and \textbf{b)} with $50\%$ environment labels, the worst-group accuracy of ISR-Cov becomes $80.9\%$, surpassing the original classifier by $18.0\%$. The compelling results demonstrate another advantage of our ISRs, the \textit{efficient utilization of environment labels}, which indicates that ISRs can be useful to many real-world datasets with only partial environment labels. \looseness=-1

\subsubsection{Applying ISRs to Pretrained Feature Extractors}
It is recently observed that CLIP-pretrained models \citep{CLIP} have impressive OOD generalization ability across various scenarios \citep{miller2021accuracy,wortsman2022robust,kumar2022finetuning}. Also, \citet{kumar2022finetuning} shows that over a wide range of OOD benchmarks, linear probing (i.e., re-training the last linear layer only) could obtain better OOD generalization performance than finetuning all parameters for CLIP-pretrained models. Notice that ISR-Mean \& ISR-Cov also re-train last linear layers on top of provided feature extractors, thus our ISRs can be used as substitutes for linear probing on CLIP-pretrained models. We empirically compare ISR-Mean/Cov vs. linear probing for a CLIP-pretrained vision transformer (ViT-B/32) in the Waterbirds dataset. As Table \ref{tab:CLIP} shows, ISRs outperform linear probing in terms of both average and worst-group accuracy, and the improvement that ISR-Mean obtains is more significant than that of ISR-Cov. This experiment indicates that our ISRs could be useful post-processing tools for deep learning practitioners who frequently use modern pre-trained (foundation) models \citep{foundation-models}.

\subsection{Multi-Class Classification}\label{sec:exp-multi-classification}

We now present the empirical results of ISR-Multiclass for multi-class classification. First, we show improvements on a synthetic multi-class linear unit test followed by the Multi-Class Colored MNIST dataset. 

\begin{figure*}[t!]
  \centering
  \includegraphics[width=\linewidth]{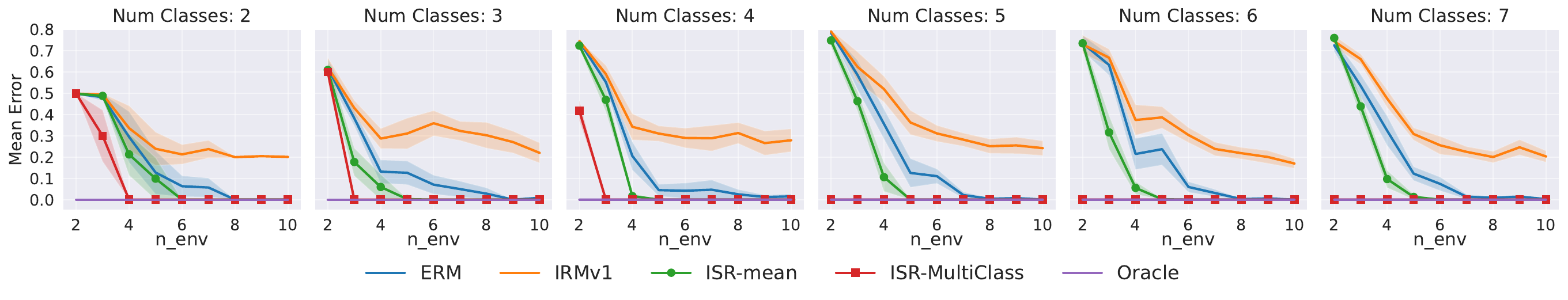}
 \label{fig:linear-unit-tests-multiclass}
  \caption{Evaluation on Multiclass Linear Unit Test Example 3s, where the y-axis denotes mean error over the test set. We have fixed $d_s = 5$ and $d_c = 5$, while $k$ varies from $2$ to $7$. As indicated by our theoretical claim: ISR-Multiclass recovers features roughly in $\lceil d_s/k \rceil + 1$ environments. For example, when $k = 3$, we achieve optimal error in $5/3 + 1 \approx 3$ environments. Similarly, beyond $k=5$, we achieve optimality in $2$ environments itself.}
\end{figure*}

\subsubsection{Multi-Class Linear Unit Tests}

We construct a multi-class version of Example 3 (and its scrambled version, Example 3s) as used in \cite{aubin2021linear}. This is a synthetic dataset based on our causal model in section \ref{sec:setup-multiclass}. Specific details of our construction can be found in Appendix \ref{supp:exp:synthetic-setup}. 

\paragraph{Evaluation Procedures}
For empirical evaluation, $R \in \mathbb{R}^{d \times d}$ is an orthonormal matrix to consider a harder version of the problem where the inputs are scrambled. The sampling of means $\mu_k, \mu_{ke}$ is done from a uniform distribution between $[0,1)$. $\nu_{inv}$ and $\nu_{spu}$ are the scale of invariant and spurious features. They are set as $\mu_{inv} = 0.1$ and  $\mu_{spu} = 1$ as regularization may encourage learning spurious features, making it harder to learn the invariant features. This is similar to Example2 in~\citep{aubin2021linear}. Further, $\sigma_c = 0.1, \sigma_e = 0.1, d_s = 5, d_c = 5$. 10,000 are sampled points per environment. $k$ varies from $2$ to $7$. 

\paragraph{Empirical Comparisons} Figure \ref{fig:linear-unit-tests} depicts the improvement of ISR-Multiclass compared to the original ISR-Mean \footnote{Note that we condition on a fixed class (0) to enable this comparison.} and ERM \cite{vapnik1992principles}. Evaluation is performed on the test split where the spurious dimensions are randomized. The Oracle is trained on this test split. From figure \ref{fig:linear-unit-tests}, we observe that ISR-Multiclass is indeed able to leverage class information and recover invariant features to achieve optimal error with the number of environments inversely proportional to $k$, confirmed by our theoretical claim. Especially in the last three plots  - with greater classes and lesser environments ($k > 5$ and $n_{env} < 5$), both ISR-Mean and ERM incur \textit{higher} error, but ISR-Multiclass takes advantage of multiple classes to \textit{improve} its performance instead and match the oracle.

\subsubsection{Multi-Class Colored MNIST}

\begin{figure*}[ht]
  \centering
  \includegraphics[width=\linewidth]{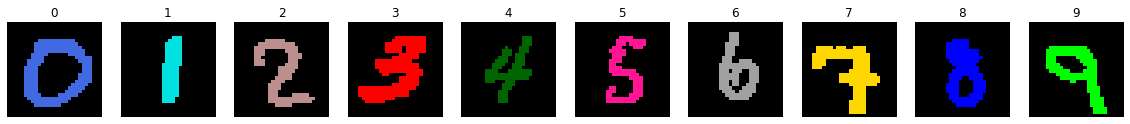}
    \label{fig:multiclass-cmnist}
  \caption{Multiclass Colored MNIST dataset.}
\end{figure*}

We consider the 10-class classification task of Colored MNIST as proposed in \cite{ahuja2021invariance}, which is a semi-synthetic dataset encoding strong spurious correlations between the digit label and color. In the train environments, every digit is highly correlated with a specific color, as depicted in Figure \ref{fig:multiclass-cmnist}. This correlation breaks down in the test environment, i.e., every digit is randomly colored.

\paragraph{Evaluation Procedures} Performance is evaluated on every group, which denotes a specific combination of $(y, color) \in \mathcal G =  \mathcal Y \times \mathcal E$. Note that there are  $10 \times 10 = 100$ groups on this dataset. During training, the samples across input groups are imbalanced owing to the spurious correlation where every digit  majorly occurs in its associated color. During test, samples across groups become balanced - thus testing a method's ability to generalize to minority groups existing in the training set. We report the average accuracy (across all groups), worst group accuracy, and worst-10 group accuracy (average across 10 worst groups). We compare the performance of ISR-Multiclass with ERM, IB-ERM, IRM \footnote{To ensure a fair comparison, IRM was re-trained with the groups denoted by digit color.} and IB-IRM which are proposed by \cite{IRM} and \cite{ahuja2021invariance}. The oracle on this dataset achieves 99.03 $\pm$ 0.08 average accuracy as per \cite{ahuja2021invariance}. More details can be found in Appendix \ref{supp:exp}. 

\paragraph{Results}

\begin{table*}[t]
\centering
    \resizebox{.99\textwidth}{!}{%
    \centering  
    \setlength{\tabcolsep}{4.5pt}
\begin{tabular}{@{}clcccccc@{}}
\toprule
\multicolumn{1}{c}{\textbf{Algorithm}} &
\multicolumn{2}{c}{\textbf{Average Accuracy}} & 
\multicolumn{2}{c}{\textbf{Worst-Group Accuracy}} &
\multicolumn{2}{c}{\textbf{Worst-10 Group Accuracy}} \\  \cmidrule(l){2-3} 
\cmidrule(l){4-5}
\cmidrule(l){6-7}

\multicolumn{1}{c}{\text{~}}&
\textbf{\small Original}& \textbf{\small ISR-Multiclass}&  \textbf{\small Original}& \textbf{\small  ISR-Multiclass}& 
\textbf{\small Original}& \textbf{\small  ISR-Multiclass} \\
\midrule
ERM & 58.20 $\pm$ 1.03 & \textbf{78.50 $\pm$ 0.76} ~&~ 0.00 $\pm$ 0.00 & \textbf{21.93 $\pm$ 13.40 } ~&~ 2.35 $\pm$ 0.59 & \textbf{39.60 $\pm$ 7.90}  ~&~\\

IB-ERM & 70.58 $\pm$ 1.24 & \textbf{81.40 $\pm$ 1.15 } ~&~ 0.94 $\pm$ 1.07 & \textbf{27.36 $\pm$ 11.42 } ~&~ 10.06 $\pm$ 2.66 & \textbf{42.63 $\pm$ 6.77}  ~&~\\

IRM & 73.85 $\pm$ 0.79 & \textbf{82.01 $\pm$ 0.97 } ~&~ 8.31 $\pm$ 2.55 & \textbf{34.33 $\pm$ 9.27 } ~&~ 25.66 $\pm$ 3.14 & \textbf{45.36 $\pm$ 6.46 }  ~&~\\

IB-IRM & 77.81 $\pm$ 0.84 & \textbf{82.95 $\pm$ 2.42 } ~&~ 9.73 $\pm$ 5.20 & \textbf{32.29 $\pm$ 6.66 } ~&~
32.14 $\pm$ 2.48 & \textbf{49.17 $\pm$ 5.63 }  ~&~\\

\bottomrule
\end{tabular}}
\caption{Evaluation of ISR Multiclass on MC-CMNIST. We report the test accuracy ($\%$) with standard deviation over 5 random trials. A value in bold indicates higher accuracy. ISR-Multiclass outperforms both average and worst group accuracies, especially for ERM. Note that the variance for worst group accuracies is high because of the less number of samples per group ($\approx 300$).} 
    \label{tab:mc-cmnist-result}

\end{table*}

Table \ref{tab:mc-cmnist-result} presents the results on MC-CMNIST.It is evident that post-processing with ISR-Multiclass significantly improves both the average and worst-group accuracies, especially prevalent for ERM and IB-ERM. While IRM and IB-IRM perform better than their ERM counterparts, ISR-Multiclass still improves the accuracy by $\approx 5-10\%$.  

\subsection{Regression}\label{sec:exp-regression}
We now study the empirical improvements of ISR-Regression across synthetic and real-life datasets, specifically on a regression Linear Unit Test, Rotated Colored Fashion MNIST, and the Law School dataset. 

\subsubsection{Regression Linear Unit Test}
We construct a linear unit test based to simulate our data generative model in section \ref{sec:setup-regression}. Specific details of the construction can be found in the appendix \ref{supp:exp:synthetic-setup}. 

\paragraph{Implementation} Here, $R \in \bR^{d \times d}$ is sampled as an orthonormal matrix for transforming $x$. The invariant mean $\mu_c$ is fixed to the vector $\mathrm{1}^{d_c}$ across all environments. The scale of invariant features $\nu_{inv} = 1$ while $\nu_{spu} = 50$ as this makes it harder to learn invariant features under regularization, motivated by a similar argument in~\citet{aubin2021linear}. $\sigma_c$ is set to $0.1$, and 10,000 data points are sampled per environment. The dimensionality $d_c$ is fixed to 5 while $d_s$ is varied from 3 to 6.

\paragraph{Evaluation Procedures} For all the methods which will be presented, hyper-parameter search is performed over $20$ data seeds and $5$ model trials. During training, Adam~\citep{kingma2014adam} is used as the optimizer and the lowest mean validation error across environments is used for model selection.

\begin{figure}[ht]
    \centering  
    \includegraphics[width=\textwidth]{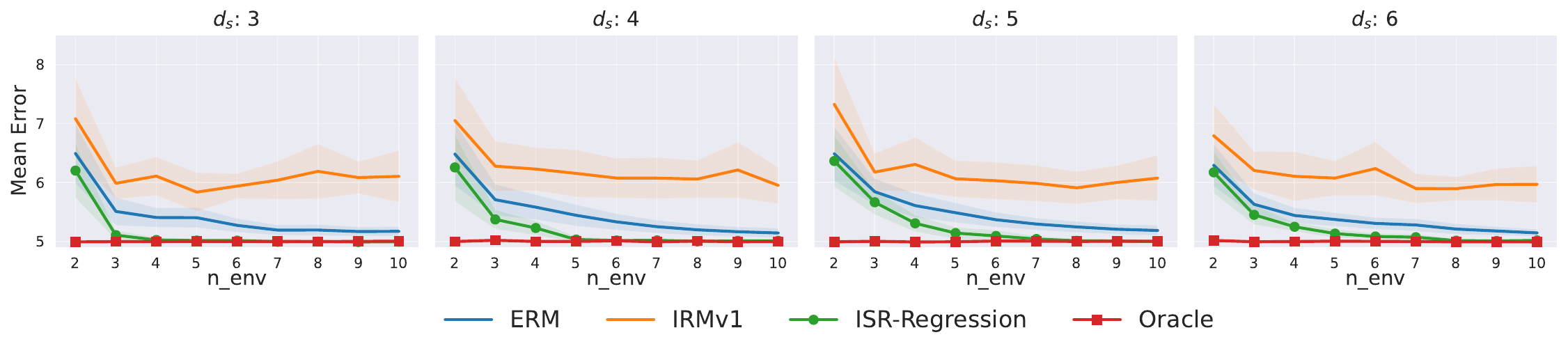}
    
    \caption{Evaluation on Regression Linear Unit Test, where the y-axis denotes mean error over the test set. The invariant dimension is fixed as $d_c = 5$ and the spurious dimensionality $d_s$ varies from $3$ to $6$ (top left to bottom right). As indicated by the theoretical claim: ISR-Regression recovers invariant features roughly in $d_s + 1$ environments, matching the performance of the Oracle beyond this number. For example, when $d_s = 3$, optimal performance is achieved in $3 + 1 = 4$ environments.}
    \label{fig:lut-regression}
\end{figure}

Figure \ref{fig:lut-regression} presents the empirical evaluation on the linear unit test for regression. Here, the oracle is trained on the test split, where the spurious dimensions are randomized. 

\paragraph{Results} It can be observed that as the dimensionality of the spurious features is increased, the performance of ISR-Regression accordingly matches the environment complexity as proved in \Cref{thm:isr-regression}. Beyond $E \geq d_s + 1$, ISR-Regression matches the oracle and achieves optimal error implying successful recovery of the invariant-feature subspace. Note that when the spurious features are greater in number i.e.\ $d_s \geq d_c$, it may be difficult to match the oracle exactly at this inflection point, possibly due to greater scale and noise in the input corresponding to the spurious dimensions ($\nu_{spu} > \nu_{inv})$. Next, even in the regime of $E < d_s$, ISR-Regression achieves lower error than both ERM and IRMv1. It is interesting to note that IRMv1 performs even worse ERM, which is possible as invariant penalties based on training error (like in IRMv1) may learn solutions relying on spurious features.

\subsubsection{Rotated Colored Fashion MNIST}
\label{sec:RCFMNIST}

\begin{figure}[ht]
    \centering
    \includegraphics[width=\linewidth]{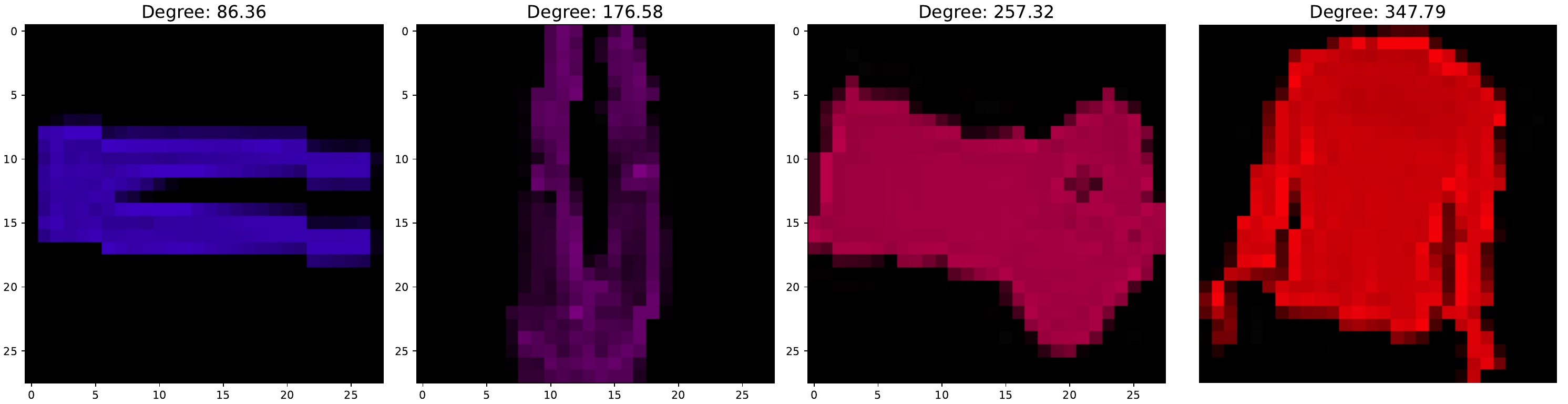}
    \caption{Rotated Colored Fashion MNIST dataset. The rotation of the object increases on moving from left to right. In the training set, a higher degree implies greater amount of red. This trend does not hold during testing. Note that the degree of this spurious correlation is set to $0.8$.}
    \label{fig:rcfmnist-dataset}
\end{figure}

Next, let us consider the Rotated Colored Fashion MNIST dataset, or RCFMNIST in short. This is a semi-synthetic dataset proposed by~\citet{cmixup} which encodes strong spurious correlations between rotation and color. Formally, the task is to predict the degree of rotation of the object i.e.\ $y$ represents the angle of rotation between $0$ and $360$ degrees. The color of the object is spuriously correlated with the degree of rotation: during training, higher the degree, greater is the amount of red in the image. However, during test time, this correlation does not exist - degree is no more correlated with the color. Therefore, any learner relying on the spurious color attribute will perform poorly on the test set, especially on groups which are present in minority during training (where the spurious correlation does not hold). Note that this is a slightly modified version from the setting considered in~\citet{cmixup} where there exists a \textit{reverse} spurious correlation on test.  Figure \ref{fig:rcfmnist-dataset} depicts this concept via a sample of the training set.

\paragraph{Obtaining Discrete Environments} It should be noted that the color varies continuously with the target variable (angle of rotation) while ISR-Regression and IRM require a set of discrete environments. To address this, the environments were formed by obtaining the \textit{mean red pixel value} and grouping images based on this value into groups of 10.

\paragraph{Implementation} A pretrained ResNet18~\citep{resnet} is finetuned on this dataset, producing a representation of dimension $512$ over which ISR is applied. For training, a batch size of $64$ is used for ERM, MixUp and C-MixUp. For IRMv1, the batch size is increased to $128$ to ensure all groups are represented in a given batch. The model is trained for $30$ epochs using the Adam~\citep{kingma2014adam} optimizer with a learning rate of $7e^{-5}$. Hyper-parameters and implementation for ERM, MixUp and C-MixUp are chosen as suggested originally by~\citet{cmixup}  \footnote{\url{https://github.com/huaxiuyao/C-Mixup}} and by~\citet{aubin2021linear} for IRMv1. For tuning the hyper-parameters of ISR (number of spurious dimensions to be scaled down and the degree of scaling), performance on the validation set is used.

\paragraph{Results}

Table \ref{tab:rcfmnist-result} presents the results from the empirical evaluation of post-processing neural network representations with ISR-Regression. Note that a group here refers to the discrete environment obtained via binning of the red pixel values. The worst group would correspond to those subpopulations of the input data where the spurious correlation breaks (e.g.\ images with a higher degree in test but with a lower red value). The performance is compared for ERM, IRM, and 2 variants of MixUp as proposed by~\citet{cmixup}. MixUp is a data augmentation strategy that chooses 2 samples and adds their linear interpolation during training. C-MixUp improves on this by smartly sampling pairs that are more similar to each other w.r.t the continuous label. Note that the Oracle on this dataset obtains an average root mean squared error i.e.\ RMSE $0.112 \pm 0.011$ and a worst-group RMSE of $0.164 \pm 0.048$.










\begin{table*}[ht!]
\centering
    \resizebox{.99\textwidth}{!}{%
    \centering  
    \setlength{\tabcolsep}{4.5pt}
\begin{tabular}{@{}clcccccc@{}}
\toprule
\multicolumn{1}{c}{\textbf{Algorithm}} &
\multicolumn{2}{c}{\textbf{Average RMSE}} & 
\multicolumn{2}{c}{\textbf{Worst-Group RMSE}} \\  \cmidrule(l){2-3} 
\cmidrule(l){4-5}

\multicolumn{1}{c}{\text{~}}&
\textbf{\small Original}& \textbf{\small ISR-Regression}&  \textbf{\small Original}& \textbf{\small  ISR-Regression} \\
\midrule
ERM & 0.262 $\pm$ 0.007 & \textbf{0.247 $\pm$ 0.001} ~&~ 0.291 $\pm$ 0.012 & \textbf{0.270 $\pm$ 0.009} ~&~\\

MixUp & 0.250 $\pm$ 0.004 & \textbf{0.245 $\pm$ 0.002} ~&~ 0.268 $\pm$ 0.008 & \textbf{0.263 $\pm$ 0.005} ~&~\\

C-MixUp & 0.260 $\pm$ 0.014 & \textbf{0.248 $\pm$ 0.008} ~&~ 0.293 $\pm$ 0.038 & \textbf{0.271 $\pm$ 0.019} ~&~\\

IRM & 0.255 $\pm$ 0.006 & 0.255 $\pm$ 0.005 ~&~ 0.277 $\pm$ 0.007 & \textbf{0.273 $\pm$ 0.009} ~&~ \\


\bottomrule
\end{tabular}}
\label{tab:rcfmnist-result}
\end{table*}

It can be observed that post-processing with ISR-Regression consistently improves performance on average and for the worst-group root mean squared error i.e.\ RMSE. Further, it should be noted that this improvement is \textit{greatest} for \textit{weaker} methods such as ERM and C-MixUp, which originally have the lowest performance. For stronger methods like MixUp, the improvement observed is lesser. An interesting point to observe is that C-MixUp performs worse than MixUp on this dataset, possibly because when considering `similar' points to sample by choosing similar label values, a correlation between the color and label can be learnt during this interpolation. Finally, for IRM (which leverages the domain information) the average performance remains similar while the worst-group performance increases, throwing light on the capability of ISR-Regression to improve performance across a spectrum of learners, even when they already leverage domain information.

\subsubsection{Law School}

\begin{figure}[t!]
    \centering
    \includegraphics[width=\textwidth] {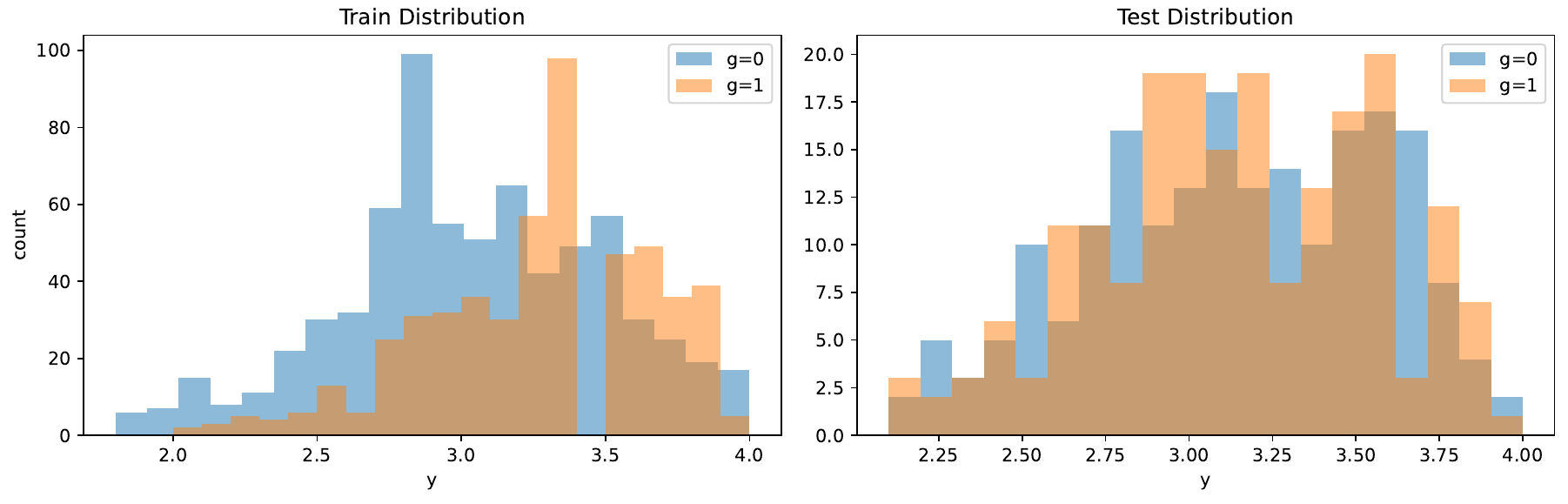}
    \caption{Law School dataset. During training (left), the target $y$ is higher for gender $g = 1$ as compared to $g = 0$. This does not hold during testing (right).}
    \label{fig:lawschool-dataset}
\end{figure}

Next, evaluation is presented on the LawSchool dataset~\citep{kearns2018preventing}. Law School is a real life tabular dataset consisting of details of people taking the bar exam along with their undergrad GPA, age, race, gender etc. For evaluation, the prediction target is the undergrad GPA or UGPA which is a continuous value between 0 and 4. The protected attribute $\mathcal A$ is considered to be the gender. In order to test for robustness to spurious correlations, the original dataset is modified such that the training set is skewed: the ratio of people with a UGPA $>3$ v/s those with UGPA $<3$ is higher for people with gender $\mathcal{A} = 1$, as compared to $\mathcal{A} = 0$. This trend does not hold during testing. Figure \ref{fig:lawschool-dataset} presents a summary of this shift.

\paragraph{Implementation}
A publicly available version \footnote{\url{https://github.com/algowatchpenn/GerryFair/blob/master/dataset/lawschool.csv}} of the LawSchool dataset has been used for evaluation, where re-sampling was done for the train, validation and test sets to simulate the spurious correlation as illustrated above. A three layer ReLU neural network was considered for this task trained with the Adam~\citep{kingma2014adam} optimizer with a learning rate of $0.001$. The network was trained for a maximum of $200$ epochs for ERM and Reweighting. For GroupDRO, the model was trained for $500$ with a weight decay of $0.0001$ and $\eta = 0.001$, based on performance on a held-out validation set. For applying ISR-Regression, PCA was first applied to the learnt neural network embeddings of size $30$ to down-sample to size $10$ and remove noisy features. Similarly, spurious dimensions were first extracted and scaled to 0. 

\paragraph{Results}
Table \ref{tab:lawschool-result} presents results for evaluation on the LawSchool dataset. For all methods, we report the R-squared metric on the fair test dataset. Note that the Oracle was trained on the test set. A group here denotes membership to a specific gender $g = 0$ or $g = 1$. 

\begin{table*}[t!]
\centering
    \resizebox{.8\textwidth}{!}{%
    \centering  
    \setlength{\tabcolsep}{4.5pt}
\begin{tabular}{@{}clcccccc@{}}
\toprule
\multicolumn{1}{c}{\textbf{Algorithm}} &
\multicolumn{2}{c}{\textbf{Average Test $R^2$}} & 
\multicolumn{2}{c}{\textbf{Worst-Group Test $R^2$}} \\  \cmidrule(l){2-3} 
\cmidrule(l){4-5}

\multicolumn{1}{c}{\text{~}}&
\textbf{\small Original}& \textbf{\small ISR-Regression}&  \textbf{\small Original}& \textbf{\small  ISR-Regression} \\
\midrule
ERM & 0.224 $\pm$ 0.019 &\textbf{0.239 $\pm$ 0.007} & 0.163 $\pm$ 0.021 & \textbf{0.193 $\pm$ 0.004}\\
Reweighting & 0.231 $\pm$ 0.012 & \textbf{0.249 $\pm$ 0.039} & 0.173 $\pm$ 0.011 & \textbf{0.212 $\pm$ 0.051}\\
GroupDRO & 0.209 $\pm$ 0.014 & \textbf{0.227 $\pm$ 0.005} & 0.172 $\pm$ 0.014 & \textbf{0.193 $\pm$ 0.001}\\
\midrule
Oracle & \multicolumn{2}{c}{0.321 $\pm$ 0.004} & \multicolumn{2}{c}{0.299 $\pm$ 0.007}\\ 
\bottomrule
\end{tabular}}
\caption{Evaluation of ISR Regression on LawSchool. The test $R^2$ is reported where a higher value denotes better performance. Note that the Oracle is trained on the test set.} 
\label{tab:lawschool-result}
\end{table*}

The ISR-Regression algorithm is used as a post-processing technique on top of neural network embeddings learnt by ERM, Reweighting and GroupDRO. In Reweighting, the group membership information is used to re-weight each sample with the weight of the corresponding group to which it belongs, thus re-weighting the overall loss to be minimized. Similarly, in GroupDRO, the reweighting happens dynamically by minimizing the worst-group loss at each step.

It can be observed that post-processing with ISR-Regression improves the test $R^2$ on an average across all groups as well as on the worst-group test set. Note that while Reweighting and GroupDRO outperform ERM on the worst-group (due to balancing out of the sample weights), ISR-Regression still improves the performance and helps close the gap of existing methods with respect to the Oracle, especially for the worst-groups on test.

\section{Conclusion}\label{sec:conclusion}
In this work, we propose ISR: a new class of algorithms for provable invariant-feature subspace recovery across the settings of classification and regression. Starting with binary classification, under a common data generative model in the literature, we propose two algorithms, ISR-Mean and ISR-Cov, to achieve domain generalization by recovering the invariant-feature subspace. We prove that ISR-Mean admits an $d_s + 1$ environment complexity and ISR-Cov obtains an $\cO(1)$ environment complexity, the minimum environment complexity that any algorithm can hope for. Furthermore, both algorithms are computationally efficient, free of local minima, and can be used off-the-shelf as a post-processing method over features learned from existing models. Next, we propose ISR-Multiclass which further improves the environment complexity to $\lceil d_s/k \rceil  + 1$ for a $k$-class classification problem, thus leveraging class information. We then present ISR-Regression for provable recovery in the setting of regression, which enjoys an environment complexity of $d_s + 1$. Empirically, we test our algorithms on synthetic and semi-synthetic benchmarks and demonstrate their superior performance when compared with other domain generalization algorithms. We also show that our proposed algorithms can be used as computationally efficient post-processing methods to increase the worst-case accuracy of (pre-)trained models by testing them on four real-world datasets spanning across image, text and tabular datasets.

\newpage
\bibliography{reference.bib}

\newpage
\appendix
\onecolumn

\section{Proof}\label{supp:proof}

\subsection{Proof of \cref{thm:isr-mean}}\label{supp:proof:isr-mean}
\begin{proof}
From \eqref{eq:def-M}, we know
\begin{align}
    \mathcal M \mathrm{\coloneqq} \begin{bmatrix}
    \bar{x}_1^\T\\
    \vdots\\
    \bar{x}_E^\T
    \end{bmatrix} \mathrm{=} \begin{bmatrix}
    \mu_c^\T A^\T \mathrm{+} \mu_1^\T B^\T\\
    \vdots\\
    \mu_c^\T A^\T \mathrm{+} \mu_E^\T B^\T\end{bmatrix}
    \mathrm{=}{\overbrace{\begin{bmatrix}
    \mu_c^\T ~~\mu_1^\T\\
    \vdots~~~~~\vdots\\
    \mu_c^\T ~~ \mu_E^\T
    \end{bmatrix}}^{\mathcal U^\T \coloneqq}} R^\T = (R\mathcal U )^\T
\end{align}
where $\mathcal U \coloneqq \begin{bmatrix}
\mu_c & \dots & \mu_c \\
\mu_1 & \dots & \mu_E 
\end{bmatrix}\in \bR^{d \times E}
$

If $E\leq d_s$, Assumption \ref{assum:non-degenerate-mean} guarantees that $\{\mu_1,\dots, \mu_E\}$ are linearly independent almost surely. Then, we have $\rank(\mathcal U) = E$. As $E > d_s$, since the first $d_c$ rows of $\mathcal U$ are the same, the rank of $\mathcal U$ is capped, i.e., $\rank(\mathcal U) = d - d_c = d_s$.

The mean-subtraction step of PCA compute the sample-mean
\begin{align}
\wt x = \frac{1}{E}\sum_{e=1}^E\bar x_e = A \mu_c + B \left(\frac{1}{E}\sum_{e=1}^E \mu_e\right) = A \mu_c + B \bar \mu,
\end{align}
where $\bar \mu \coloneqq \frac{1}{E}\sum_{e=1}^E \mu_e $, and then subtracts $\wt x^\T$ off each row of $\mathcal M$ to obtain
\begin{align}
    \wt {\mathcal M} \mathrm{\coloneqq} \begin{bmatrix}
    \bar{x}_1^\T - \wt x^\T \\
    \vdots\\
    \bar{x}_E^\T - \wt x^\T
    \end{bmatrix} \mathrm{=} \begin{bmatrix}
    (\mu_1-\bar \mu)^\T B^\T \\
    \vdots\\
    (\mu_E-\bar \mu)^\T B^\T\end{bmatrix}
    \mathrm{=}{\overbrace{\begin{bmatrix}
    \mu_1^\T - \bar \mu^\T\\
    \vdots\\
    \mu_E^\T - \bar \mu^\T
    \end{bmatrix}}^{\wt{\mathcal U}^\T \coloneqq}} B^\T = (B \wt {\mathcal U} )^\T\in \bR^{ d_s \times d}
\end{align}
where $\wt{\mathcal U} \coloneqq \begin{bmatrix}
\mu_1 - \bar \mu & \dots & \mu_E - \bar \mu 
\end{bmatrix}\in \bR^{d_s \times E}
$

Similar to the analysis of $\mathcal U$ above, we can also analyze the rank of $\wt{\mathcal U}$ in the same way. However, different from $\cU$, we have $\rank(\wt\cU) = \min\{d_s, E-1\}$, where the $-1$ comes from the constraint $\sum_{e=1}^E (\mu_e - \bar \mu) = 0$ that is put by the mean-subtraction.

Suppose $E\geq d_s +1$, then $\rank(\wt\cU) = d_s$. The next step of PCA is to eigen-decompose the sample covariance matrix
\begin{align*}
    \frac{1}{E}\wt\cM^\T \wt\cM =  \frac{1}{E} (B \wt {\mathcal U} ) (B \wt {\mathcal U} )^\T &= \frac{1}{E} B (\wt {\mathcal U} \wt {\mathcal U}^\T ) B^\T\\
    &= \frac{1}{E} 
    \begin{bmatrix}
    A&B
    \end{bmatrix}
    \begin{bmatrix}
    \textbf{0}_{d_c \times d_c} & \textbf{0}_{d_c \times d_s}\\
    \textbf{0}_{d_s\times d_c} & \wt {\mathcal U} \wt {\mathcal U}^\T
    \end{bmatrix}
    \begin{bmatrix}
    A^\T\\
    B^\T
    \end{bmatrix}\in \bR^{d \times d}\eq \label{eq:supp:proof:isr-mean:cov-eigendecom}
\end{align*}
where $\textbf{0}_{n\times m}$ is a $n\times m$ matrix with all zero entries, and $\wt {\mathcal U} \wt {\mathcal U}^\T \in \bR^{d_s \times d_s}$ is full-rank because  $\rank(\wt\cU) = d_s$. 

Combining with the fact that $R=[A B]$ is full-rank (ensured by Assumption \ref{assum:full-rank-transform}), we know that $\rank(\frac{1}{E}\wt\cM^\T \wt\cM) = d_s$. Therefore, $\frac{1}{E}\wt\cM^\T \wt\cM$ is positive-definite.

As a result, the eigen-decomposition on $\frac{1}{E}\wt\cM^\T \wt\cM$ leads to an eigen-spectrum of $d_s$ positive values and $d_c = d - d_s$ zero eigenvalues. 

Consider ascendingly ordered eigenvalues $\{\lambda_1,\dots, \lambda_d\}$, and compose a diagonal matrix $S$ with these eigenvalues as in ascending order, i.e., $S\coloneqq\mathrm{diag}(\{\lambda_1,\dots, \lambda_d\})$. Denote the eigenvectors corresponding with these eigenvalues as $\{P_1,\dots, P_{d_c}\}$, and stack their transposed matrices as
\begin{align}
    P \coloneqq
    \begin{bmatrix}
    P_1^\T \\
    \vdots \\
    P_d^\T
    \end{bmatrix}
    \in \bR^{d\times d}
\end{align}
Then, we have the equality
\begin{align}
    \frac{1}{E} B (\wt {\mathcal U} \wt {\mathcal U}^\T ) B^\T = \frac{1}{E}\wt\cM^\T \wt\cM = P S P^\T 
\end{align}

Since the first $d_c$ diagonal entries of $S$ are all zeros and the rest are all non-zero, the dimensions of $P$ that correspond to non-zero diagonal entries of $S$ can provide us with the subspace spanned by the $d_s$ spurious latent feature dimensions, thus the rest dimensions of $P$ (i.e., the ones with zero eigenvalues) correspond to the subspace spanned by $d_c$ invariant latent feature dimensions, i.e.,
\begin{align}
    \mathrm{Span}(\{P_i^\T R: i\in[d],~ S_{ii} = 0\}) = \mathrm{Span}(\{\mathbf{\hat d_c^1},\dots, \mathbf{\hat d_c^{d_c}}\})
\end{align}
Since the diagonal entries of $S$ are sorted in ascending order, we can equivalently write it as
\begin{align}\label{eq:thm:inv-subspace-recovery:span-equality:supp}
    \mathrm{Span}(\{P_1^\T R, \dots, P_{d_c}^\T R\}) = \mathrm{Span}(\{\mathbf{\hat d_c^1},\dots, \mathbf{\hat d_c^{d_c}}\})
\end{align}
Then, by \cref{prop:optimal-inv-pred} (i.e., Definition 1 of \citet{risks-of-IRM}) and Lemma F.2 of \citet{risks-of-IRM}, for the ERM predictor fitted to all data that are projected to the recovered subspace, we know it is guaranteed to be the optimal invariant predictor (defined in \cref{prop:optimal-inv-pred} as Eq. \eqref{eq:optimal-inv-pred}).

\textbf{Case $E \leq d_s$}
Now, let us consider the case when $E \leq d_s$, $\text{rank}(\widehat{U}) = \min(E-1, d_s) =  E - 1$. The eigen-decomposition of $\wt\cM^\T \wt\cM$ will yield $E - 1$ strictly positive eigenvalues and $d - (E - 1) > d_c$ zero eigenvalues. In this case,
\begin{align}
\label{eq:superset}
    \text{Span}(\{P_i^\T R: P_i \in P_{zero}\}) \supset \mathrm{Span}(\{\mathbf{\hat d_c^1},\dots, \mathbf{\hat d_c^{d_c}}\})
\end{align} 
where $P_{zero}$ belongs to the set of eigenvectors corresponding to zero eigenvalues. 
One could alternately leverage the transformation matrix obtained by stacking eigenvectors in $P_{positive}$ (set of eigenvectors corresponding to strictly positive eigenvalues)i.e.\ $[P_1,\cdots,P_{E-1}]^\T = P^{''} \in \mathbb{R}^{(E-1) \times d}$ to \textit{partially} recover the spurious feature subspace:
\begin{align}
\label{eq:subset}
    \text{Span}(\{P_i^\T R: P_i \in P_{positive}\}) \subset \mathrm{Span}(\{\mathbf{\hat d_s^1},\dots, \mathbf{\hat d_s^{d_s}}\})
\end{align}
Following this, the nullspace of $P^{''}$ can be used to obtain $Nullspace(P^{''}) = P^{'} \in \mathbb{R}^{(d-(E-1)) \times d}$ to recover a \textit{partial} transformation to the invariant-feature subspace under the available information. While both of these present two ways to recover the invariant-feature subspace, it is suggested to use equation \eqref{eq:subset}, which only removes those spurious dimensions that \textit{significantly vary} across environments.

\end{proof}

\subsection{Proof of \cref{thm:isr-cov}}\label{supp:proof:isr-cov}
\begin{proof}
From \eqref{eq:delta-sigma-expression}, we know
\begin{align}
\Delta \Sigma \coloneqq \Sigma_{e_1} - \Sigma_{e_2} = (\sigma_{e_1}^2 - \sigma_{e_2}^2)BB^\T \in \bR^{d \times d}
\end{align}
Assumption \ref{assum:non-degenerate-cov} guarantees that $\sigma_{e_1}^2 - \sigma_{e_2}^2 \neq 0$, and Assumption \ref{assum:full-rank-transform} ensures that $\rank(B) = d_s$. Thus, eigen-decomposition on $\Delta \Sigma$ leads to exactly $d_c$ zero eigenvalues and $d_s = 1- d_c$ non-zero eigenvalues.
One just need to follow the same steps as \eqref{eq:supp:proof:isr-mean:cov-eigendecom}-\eqref{eq:thm:inv-subspace-recovery:span-equality:supp} to finish the proof.
\end{proof}

\subsection{Proof of \cref{thm:isr-multiclass}}\label{supp:proof:isr-multiclass}

\begin{proof} Consider the matrix $\mathcal{M}_{total}$ as per \eqref{eq:def-Mtotal}:
\begin{align}
    \mathcal{M}_{total} \mathrm{\coloneqq} \begin{bmatrix}
    P_{1} | P_{2} | \cdots | P_{k}
    \end{bmatrix} \in \mathbb{R}^{d \times (E-1)k}
\end{align}

By definition, $\text{rank}(\mathcal{M}_{total}) \leq \min(d, (E - 1) \times k)$, which trivially implies:
\begin{align}
    \text{rank}( \mathcal{M}_{total}) \leq (E - 1) \times k
\end{align} 

Recall that each $P_k$ recovers the spurious dimension specific to class $k$. In order to recover the underlying $d_s$ dimensional subspace, the rank of $\mathcal{M}_{total} = d_s$. Combining this fact with the above statement, the following inequality is obtained:
\begin{align}
    E - 1 \geq d_s / k \\
    \label{eq:ineq-multiclass}
    E \geq d_s/k + 1
\end{align}

Thus, the minimum number of environments required to recover the spurious (thus invariant) feature subspace \textit{benefits} by leveraging information from classes. The greater the number of classes, the lesser the number of training environments we require to recover the $d_c$ dimensional invariant features. It should be noted that this decrease is observed while only leveraging the $1^{st}$ order moments of the class conditional data distribution. 

Assuming condition \ref{assum:non-degenerate-mean-multiclass} is satisfied,  the rank of $\mathcal{M}_{total}$ is capped at $d_s$. Thus, the SVD will lead to $d_s$ strictly positive singular values. Then, one can obtain the $d_s$ eigenvectors corresponding to these singular values, which span the spurious dimensions. The transformation matrix will be $P'\in \bR^{d_s\times d}$. Since $z_s \perp z_c$ as per the setup \ref{sec:setup-multiclass}, the null space of $P'$ will correspond to vectors spanning the $d-d_s = d_c$ dimensions as follows:
\begin{align}
    NullSpace(P') = P'' \in \mathbb{R}^{d_c\times d}
\end{align}

Finally, training on this invariant subspace helps obtain the optimal invariant predictor as per \ref{prop:optimal-inv-pred}, which completes the proof.
\end{proof}

\subsection{Proof of \cref{thm:isr-regression}}\label{supp:proof:isr-regression}
\begin{proof} Consider the matrix $\cM$ as per \eqref{eq:def-M-regression}:
\begin{align}
    \mathcal M \mathrm{\coloneqq} \begin{bmatrix}
    \bar{x}_1^\T\\
    \vdots\\
    \bar{x}_E^\T
    \end{bmatrix} \in \bR^{E \times d}
\end{align}

Recall that under the infinite sample setting, each row $\bar{x}_e^\top$ of $\cM$ can be represented as the following mean estimate:
\begin{align}
    \bar{x_e} = A\mu_c + B\mu_e
\end{align}
Thus, $\cM$ can now be expressed as:
\begin{align}
    \mathcal M \mathrm{\coloneqq} \begin{bmatrix}
    \bar{x}_1^\T\\
    \vdots\\
    \bar{x}_E^\T
    \end{bmatrix} 
    \mathrm{=} \begin{bmatrix}
    \mu_c^\T A^\T \mathrm{+} \mu_1^\T B^\T\\
    \vdots\\
    \mu_c^\T A^\T \mathrm{+} \mu_E^\T B^\T\end{bmatrix}
    \mathrm{=}{\overbrace{\begin{bmatrix}
    \mu_c^\T ~~\mu_1^\T\\
    \vdots~~~~~\vdots\\
    \mu_c^\T ~~ \mu_E^\T
    \end{bmatrix}}^{U^\T \coloneqq}} R^\T
\end{align}
where 
\begin{align}
    U = \begin{bmatrix}
    \mu_c ~~\mu_c~~\cdots~~\mu_c \\
    \mu_1 ~~\mu_2~~\cdots~~\mu_E \\
    \end{bmatrix} \in \bR^{d \times E}
\end{align}

Note that once we have this formulation, one can follow the same steps as the proof in \cref{supp:proof:isr-mean}.
which completes the proof.

\end{proof}

\section{Experimental Details}\label{supp:exp}

\subsection{Setups of Synthetic Datasets}\label{supp:exp:synthetic-setup}

\paragraph{Example-2}
This is a binary classification task that imitate the following example inspired by \citet{IRM, beery2018recognition}: while most cows appear in grasslands and most camels appear in desserts, with small probability such relationship can be flipped. In this example, \citet{aubin2021linear} define the animals as invariant features with mean $\pm \mu_c$ and the backgrounds as spurious features with mean $\pm \mu_e$. \citet{aubin2021linear} also scale the invariant and spurious features with $\nu_c$ and $\nu_e$ respectively. To be specific, we set $\mu_{c} = \mathbf{1}_{d_c}$ (i.e., a $d_c$-dimensional vector with all elements equal to $1$) , $\mu_{e} = \mathbf{1}_{d_e}$, $\nu_c = 0.02$ and $\nu_e = 1$. For any training environment $e \in \mathcal{E}$, \citet{aubin2021linear} construct its dataset $\mathcal D_{e}$ by generating each input-label pair $(x,y)$ in the following process:
\begin{align*}
    j_e &\sim \text{Categorical}\left(p^e s^e , (1 - p^e) s^e, p^e (1-s^e) , (1 - p^e) (1-s^e)\right)\\
        z_c &\sim
        \left\{\begin{array}{lr}
            +1\cdot \left(\mu_c+\mathcal{N}_{d_c}(0, 0.1) \right) \cdot \nu_c & \text{ if } j_e \in \{1, 2\},\\
            -1\cdot \left(\mu_c + \mathcal{N}_{d_c}(0, 0.1)\right) \cdot \nu_c & \text{ if } j_e \in \{3, 4\},\\
        \end{array}\right.\\
        z_e &\sim
        \left\{\begin{array}{lr}
            +1\cdot \left(\mu_e + \mathcal{N}_{d_s}(0, 0.1) \right) \cdot \nu_e & \text{ if } j_e \in \{1, 4\},\\
            -1\cdot \left(\mu_e + \mathcal{N}_{d_s}(0, 0.1)  \right)  \cdot \nu_e & \text{ if } j_e \in \{2, 3\},
        \end{array}\right. 
\end{align*}
\begin{align*}
        z \leftarrow \begin{bmatrix}
        z_c\\
        z_e
        \end{bmatrix}, \qquad 
        y \leftarrow
        \left\{\begin{array}{lr}
        1 & \text{ if } 1_{d_c}^\T z_c > 0,\\
        0 & \text{else}
        \end{array}\right. ,\qquad 
     x = Rz \quad 
     \text{with} \quad
     R = I_{d} , 
\end{align*}
where the background probabilities are $p^{e=0} = 0.95$, $p^{e=1} = 0.97$, $p^{e=2} = 0.99$ and the animal probabilities are $s^{e=0} = 0.3$, $s^{e=1} = 0.5$, $s^{e=2} = 0.7$. If there are more than three environments, the extra environment variables are drawn according to $p^{e} \sim \textrm{Unif}(0.9, 1)$ and $s^{e} \sim \textrm{Unif}(0.3, 0.7)$.

\paragraph{Example-3}~
This is a linear version of the spiral binary classification problem proposed by \citet{parascandolo2020learning}. In this example, \citet{aubin2021linear} assign the first $d_c$ dimensions of the features with an invariant, small-margin linear decision boundary, and the reset $d_e$ dimensions have a changing, large-margin linear decision boundary. To be specific, for all environments, the $d_c$ invariant features are sampled from a distribution with a constant mean, while the means are sampled from a Gaussian distribution for the $d_e$ spurious features. In practice set $\gamma = 0.1 \cdot \mathbf{1}_{d_c}$, $\mu_e \sim \mathcal{N}(\mathbf{0}_{d_c}, I_{d_c})$, and $\sigma_c=\sigma_e=0.1$, for all environments. For any training environment $e \in \mathcal{E}$, \citet{aubin2021linear} construct its dataset $\mathcal D_{e}$ by generating each input-label pair $(x,y)$ in the following process:
\begin{align*}
    y &\sim \text{Bernoulli}\left(\frac{1}{2}\right),\\
    z_c &\sim
    \left\{\begin{array}{lr}
        \mathcal{N}(+\gamma, \sigma_c I_{d_c}) & \text{ if } y = 0,\\
        \mathcal{N}(-\gamma, \sigma_c I_{d_c}) & \text{ if } y = 1;\\
    \end{array}\right.\\
    z_e &\sim
    \left\{\begin{array}{lr}
        \mathcal{N}(+\mu_e, \sigma_e I_{d_s}) & \text{ if } y = 0,\\
        \mathcal{N}(-\mu_e, \sigma_e I_{d_s}) & \text{ if } y = 1;\\
    \end{array}\right.
\end{align*}
\begin{align*}
        z \leftarrow \begin{bmatrix}
        z_c\\
        z_e
        \end{bmatrix}, \qquad 
     x = Rz \quad 
     \text{with} \quad
     R = I_{d} 
\end{align*}

\paragraph{Example-3'} As explained in \cref{sec:lut}, in order to make Example-3 follow Assumption \ref{assum:non-degenerate-cov}, we slightly modify the variance of the  features in Example-3 so that $\sigma_c=0.1$ and $\sigma_e\sim \mathrm{Unif}(0.1, 0.3)$. All the rest settings are unchanged.

\paragraph{Example-2s/3s/3s'} In order to increase the difficulty of the tasks, we defined the ``scrambled`` variations of the three problems described above. To build the scrambled variations, we no longer use the identity matrix $I_d$ as the transformation matrix $R$; instead, a random orthonormal matrix $R\in\mathbb{R}^{d\times d}$ is applied to the features for all environments $e\in\mathcal{E}$. The random transformation matrix is built from a Gaussian matrix (see the code \url{https://github.com/facebookresearch/InvarianceUnitTests} of \citet{aubin2021linear} for details).

\paragraph{Multiclass Linear Unit Test}
In this dataset, the target $y$ is sampled from a multinomial distribution of uniform probability $1/k$, where k is the number of classes. Then, the first $d_c$ invariant features are sampled from a Gaussian distribution where the mean depends on the class label. Similarly, the next $d_s$ spurious features are sampled from a Gaussian distribution where the mean now depends on the class label \textit{as well as} the environment label. This can be formulated as follows:

For a given environment $e$, 
\begin{align}
    y &\sim \text{Multinomial}\left(\frac{1}{k}\right),\\
    z_c &\sim
    \left\{\begin{array}{lr}
        \mathcal{N}(\mu_k, \sigma_c I_{d_c}) * \nu_{inv} & \text{ for } y = k,\\
    \end{array}\right.\\
    z_e &\sim
    \left\{\begin{array}{lr}
        \mathcal{N}(\mu_{ke}, \sigma_e I_{d_s}) * \nu_{spu} & \text{ for } y = k, \text{ env} = e,\\
    \end{array}\right.\\
        z &\leftarrow \begin{bmatrix}
        z_c\\
        z_e
        \end{bmatrix}, \qquad 
     x = Rz
\end{align}

\paragraph{Regression Linear Unit Test}
The dataset is generated as follows. For a given environment $e$, 
\begin{align}
    z_c &\sim
    \left\{\begin{array}{lr}
        \mathcal{N}(\mu_c, \sigma_c I_{d_c}) * \nu_{inv},\\
    \end{array}\right.\\
    y &\sim w_c ^\top z_c + b_c,\\
    z_e &\sim (W_{cs}^e z_c + b_e) * \nu_{spu}\\
        z &\leftarrow \begin{bmatrix}
        z_c\\
        z_e
        \end{bmatrix}, \qquad 
     x = Rz
\end{align}

\subsection{Experiments on Synthetic Datasets}

\paragraph{Code} We adopt the codebase of Linear Unit-Tests \citep{aubin2021linear}, which provide implementations of Example-2/2s/3/3s and multiple algorithms (including IRMv1, IGA, ERM, Oracle). This codebase is released at \url{https://github.com/facebookresearch/InvarianceUnitTests}.

\paragraph{Hyper-parameters} Similar to \citet{aubin2021linear}, we perform a hyper-parameter search of 20 trials. For each trial, we train the algorithms on the training split of all environments for 10K full-batch Adam \citep{adam} iterations. We run the search for ISR-Mean and ISR-Cov algorithms on all examples, and run the search for ERM, IGA \citep{koyama2020out}, IRMv1 \citep{IRM} and Oracle on Example-3' and Example-3s'. We choose the hyper-parameters that minimize the mean error over the validation split of all environments. The experiment results for ERM, IGA, IRMv1 and Oracle on Example-2, Example-2s, Example-3 and Example-3s are from \citep{aubin2021linear}, thus we do not perform any search on them.

For Multiclass Linear Unit Tests, in our experiments, $\sigma_c = 0.1, \sigma_e = 0.1, d_s = 5, d_c = 5$. We sample 10,000 points per environment. $k$ varies from $2$ to $7$. For all methods, we perform a hyperparameter search over $5$ data seeds and $5$ model trials. In every trial, we train the algorithm on the train split and use the Adam \cite{kingma2014adam} optimizer for optimization. The model with the least mean validation error across all environments is chosen.

\paragraph{Multi-Class Colored MNIST}

In this dataset, for every digit, the corresponding color is highly correlated in the training set. This correlation breaks during testing. 

We directly employ the Multiclass Colored MNIST dataset, models and hyperparameters provided by \cite{ahuja2021invariance} at \url{https://github.com/ahujak/IB-IRM}.  For ERM, IB-ERM, IRM and IB-IRM, we run a sweep over hyperparameters using the grid as suggested above. The best model is chosen by using train domain validation (\cite{gulrajani2020search}). ISR-Multiclass is applied on the last-layer over the classification weights to enable the invariant-feature subspace transformation. Note that ISR-Multiclass uses color labels as the group information, and we ensure this same definition applies to IRM to ensure a fair comparison.

\subsection{Experiments on Real Datasets}
\paragraph{Training} We directly use models, hyper-parameters and running scripts provided by authors of \citet{sagawa2019distributionally} in \url{https://github.com/kohpangwei/group_DRO}. Specifically, they use ResNets \citep{resnet} for Waterbirds and CelebA, and deploy BERT \citep{bert} for MultiNLI. We train the neural nets following the official running scripts\footnote{Provided in \url{https://worksheets.codalab.org/worksheets/0x621811fe446b49bb818293bae2ef88c0}}. over 10 random seeds for Waterbirds/CelebA/MultiNLI. Each run leads to one trained neural network selected on the epoch with the highest worst-group validation accuracy.

\paragraph{ISR-Mean} There are only $E=2$ environments for Waterbirds, CelebA and MultiNLI and ISR-Mean can only identify a $\min\{E-1, d_s\}$-dimensional spurious subspace. Thus we assume $d_s=1$ for the three datasets when applying ISR-Mean. 

\paragraph{ISR-Cov} For real datasets, we do not know the $d_s$ of the learned features, thus we have to treat $d_s$ as a hyperparameter for Algorithm \ref{algo:isr-cov}. 

\paragraph{Numerical Techniques}
The feature space of learned models is usually of a high dimension (e.g., 2048 for ResNet-50 in Waterbirds/CelebA), while the features of training data usually live in a subspace (approximately). Thus, we typically apply dimension reduction to features through a PCA. Then, to overcome some numerical instability challenges, we apply ISRs in an equivalent approach: we first identify the spurious-feature subspace, and then reduces scales of features along the spurious-feature subspace. The final step of fitting linear predictors in Algorithm \ref{algo:isr-mean}/\ref{algo:isr-cov} is done by logistic regression solver provided in scikit-learn \citet{sklearn}. But in some cases, we find that directly adapting the original predictor of the trained model also yields good performance. 

\end{document}